\documentclass[runningheads]{llncs}

\usepackage{eccv}

\usepackage{eccvabbrv}

\usepackage[dvipsnames]{xcolor}
\usepackage{colortbl}
\usepackage{graphicx}
\usepackage{array}
\usepackage{booktabs}
\usepackage{multirow}
\usepackage{rotating}
\usepackage{comment}
\usepackage{makecell}
\usepackage{subcaption}
\usepackage[ruled,vlined]{algorithm2e}

\usepackage[accsupp]{axessibility}  % Improves PDF readability for those with disabilities.

\usepackage[pagebackref,breaklinks,colorlinks,citecolor=eccvblue]{hyperref}
\usepackage[width=122mm,left=18mm,right=18mm, paperwidth=158mm, height=193mm,top=22mm,bottom=22mm, paperheight=237mm]{geometry}
\usepackage[font=small,skip=4pt]{caption}

\def\eqref#1{Eq.~(\ref{#1})}

\newcommand{\tightparagraph}[1]{\vspace{0.4ex}\noindent\textbf{#1}\quad}

\newcommand{\RR}{\mathbb{R}}

\begin{document}

\title{Register Any Point: Scaling 3D Point Cloud Registration by Flow Matching}

\titlerunning{Register Any Point}

\author{
Yue Pan\inst{1} \quad Tao Sun\inst{2} \quad Liyuan Zhu\inst{2} \quad Lucas Nunes\inst{3} \\ Iro Armeni\inst{2} \quad Jens Behley\inst{1} \quad  Cyrill Stachniss\inst{1}\\
}

\authorrunning{Y.~Pan et al.}

\institute{Center for Robotics, University of Bonn, Germany \and
Stanford University, USA \and
RWTH Aachen University, Germany \\
% \email{yue.pan@igg.uni-bonn.de}
% \vspace{-1em}
}

\maketitle

\begin{figure}[h]
  \vspace{-18pt}
  \centering
  \includegraphics[width=0.97\linewidth]{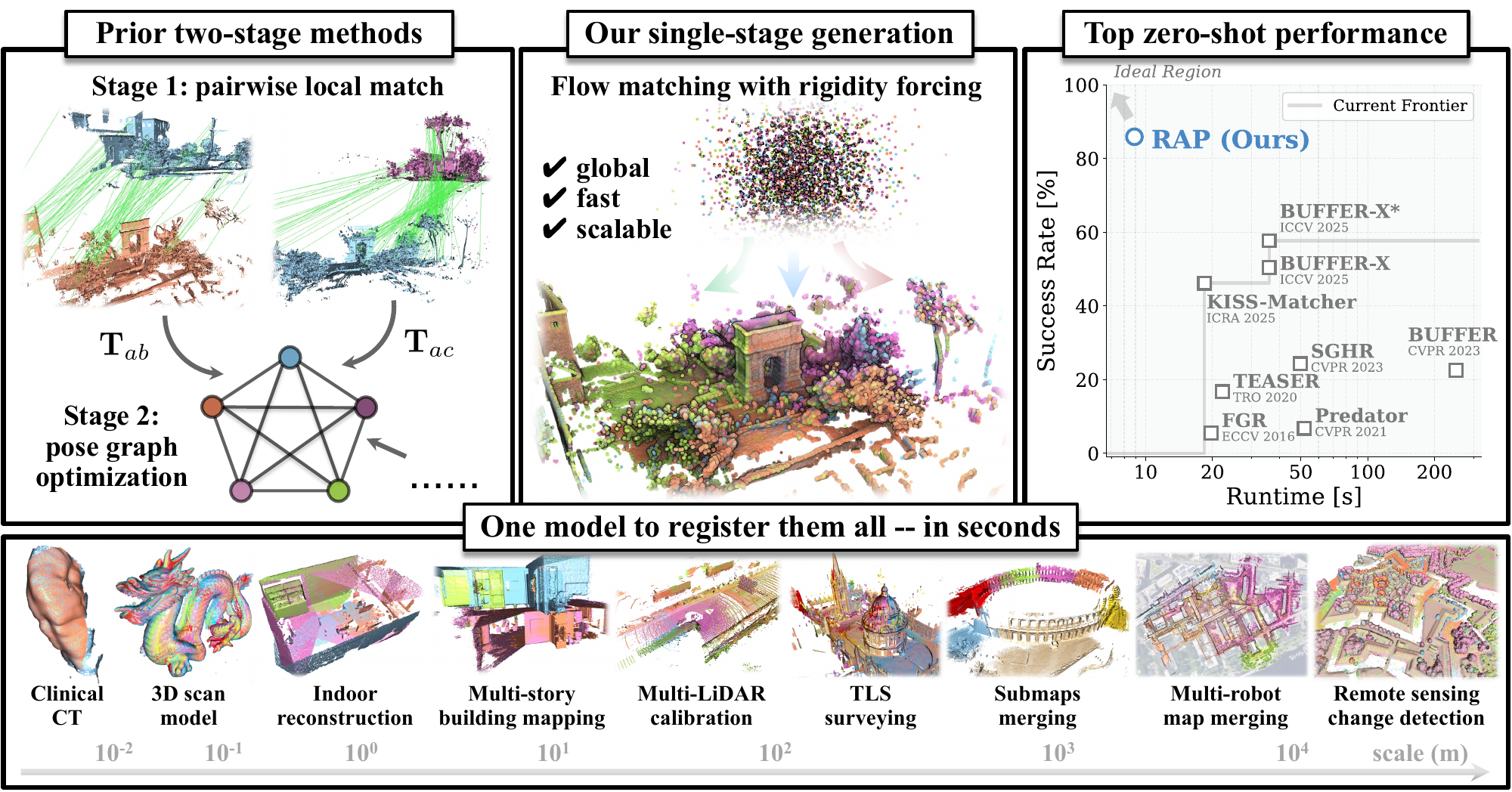}
  \caption{
    \textbf{Our method for multi-view point cloud registration.}
    Prior works perform pairwise registration via correspondence matching and then conduct pose graph optimization (top-left). We use a single-stage flow matching model to generate the registered points (top-middle). Our model generalizes across various view counts, scales, and sensor modalities (bottom) and achieves superior zero-shot performance with the shortest runtime on the cross-domain multi-view registration benchmark (top-right).
  }
  \label{fig:teaser}
  \vspace{-24pt}
\end{figure}

\begin{abstract}
  Point cloud registration aligns multiple unposed point clouds into a common reference frame and is a core step for 3D reconstruction and robot localization without initial guess.
  In this work, we cast registration as conditional generation: a learned, continuous point-wise velocity field transports noisy points to a registered scene, from which the pose of each view is recovered.
  Unlike prior methods that perform correspondence matching to estimate pairwise transformations and then optimize a pose graph for multi-view registration, our model directly generates the registered point cloud, yielding both efficiency and point-level global consistency.
  By scaling the training data and conducting test-time rigidity enforcement, our approach achieves state-of-the-art results on existing pairwise registration benchmarks and on our proposed cross-domain multi-view registration benchmark. The superior zero-shot performance on this benchmark shows that our method generalizes across view counts, scene scales, and sensor modalities even with low overlap. [\href{https://register-any-point.github.io/}{\texttt{project page}}] 
  % It further supports various downstream tasks in robotics, remote sensing, as well as architecture, engineering, and construction applications.
  % \keywords{Point Cloud Registration \and Flow Matching}
\end{abstract}

%%%%%%%%%%%%%%%%%%%%%%%%%%%%%%%%%%%%%%%%%%%%%%%%%%%%%%%%%%%%%%%%%%%%%%%%%%%%%%%%
\section{Introduction}
\label{sec:intro}

Point cloud registration is a cornerstone in 3D vision, robotics, and photogrammetry with broad applications, from merging multiple partial 3D scans into a consistent 3D model to localizing sensors in an existing 3D map for downstream tasks, including simultaneous localization and mapping (SLAM)~\cite{zhang20243d,pan2021icra-mvls}, 3D reconstruction~\cite{dong2020jprs-whutls}, and robotic manipulation~\cite{pomerleau2015ftr}. 
Yet, obtaining reliable registration in the wild without an initial guess is a hard problem. 
Real-world data is sparse, noisy, and non-uniform in density; sensors differ in modality and calibration; overlaps between point clouds can be small, and local matches can be ambiguous~\cite{huang2021cvpr-predator,sun2025jprs,an2024pointtr,liu2024cvpr-eyoc}.

Prevailing approaches for multi-view point cloud registration follow a two-stage pipeline: align all overlapping pairs of scans independently, then solve a global pose graph optimization to enforce consistency~\cite{schneider2012isprs,huang2021ral-bundle}. 
Pairwise alignment typically relies on matching local feature correspondences with a robust estimator~\cite{fischler1981cacm, yang2020tro,lim2024ijrr-quatropp, babin2019icra, laserna2025tro}. 
While conceptually appealing, this has two limitations: (i) quadratic complexity: cost scales quadratically with the number of scans due to exhaustive pairwise registration across all pairs; 
and (ii) limited global context: the pairwise stage limits capturing global context, hurting performance with low overlap and incomplete observations.
Although specialized modules can improve low-overlap pairwise registration~\cite{huang2021cvpr-predator,yao2024iccv-parenet} and some works conduct hierarchical registration~\cite{dong2018jprs} or edge selection~\cite{wang2023cvpr-sghr} to avoid the quadratic cost, these add complexity while remaining tied to iterative pose-graph refinement sensitive to pairwise alignment errors.

Recent 3D vision research departs from this two-stage pipeline by leveraging feed-forward and generative models.
In image-based 3D reconstruction, feed-forward approaches~\cite{wang2024cvpr-dust3r, leroy2024eccv} encapsulate the entire structure-from-motion process into a single neural network, directly producing globally consistent poses and dense geometry from a set of images.
VGGT~\cite{wang2025cvpr-vggt} demonstrates that a large transformer can infer all key 3D attributes, including camera poses and depth maps, from one or many views in a single pass.
In the point cloud domain, Rectified Point Flow~(RPF)~\cite{sun2025neurips} pioneered a generative approach to pose estimation by learning a continuous flow field that moves points from random noise to their assembled target positions for multiple object-centric benchmarks. 
These findings suggest that a single feed-forward model can holistically reason about multiple partial observations and produce a consistent 3D alignment, given sufficient capacity and training data.

%% Issues of the generative models

Scaling such single-stage models to large-scale, multi-view 3D registration, however, raises another key challenge: the sampling process does not always yield stable, perfectly rigid predictions, especially in cluttered environments where geometry is more diverse than in object-centric settings. 
Even with an explicit projection step of the final prediction onto $\mathrm{SE}(3)$, as in RPF~\cite{sun2025neurips}, the post-hoc correction cannot constrain the entire flow trajectory. 
Thus, the sampled flows can drift away from the flow distribution on which the model was trained, limiting performance.

This motivates our work, a scalable generative model that aligns multiple point clouds in a single stage while explicitly enforcing rigidity.
Instead of exhaustive pairwise transformation estimation, the model learns to transform all input point clouds directly into a canonical coordinate frame, effectively fusing them into a coherent scene.
To make generation robust and satisfy rigid constraints, we propose using rigidity as a guidance signal for flow sampling at test time.
To train at scale, we curate over 100k multi-view registration instances from 17 diverse datasets spanning object-centric, indoor, and outdoor settings. 
Supervising in Euclidean space across this mixture of data provides strong scene priors that enable the model to complete partial views and generalize across scales and sensor modalities.
To address the lack of a unified evaluation protocol for cross-domain generalization, we also introduce a multi-view registration benchmark spanning five scene categories across diverse scales, sensors, and overlap conditions.
We will release our code, model, and proposed benchmark to facilitate reproducibility at: \mbox{\url{https://github.com/PRBonn/RAP}}.

In summary, our contributions are three-fold:
\begin{itemize}
    \item We propose RAP, a generative flow-matching model for single-stage multi-view point cloud registration, together with a rigidity-forcing sampling strategy that enforces per-scan rigid constraints at test time, achieving state-of-the-art performance on both pairwise and multi-view point cloud registration benchmarks.
    \item We develop a large-scale training recipe that aggregates over 100k multi-view registration instances from 17 heterogeneous datasets, enabling strong generalization across diverse scenarios, scales, and sensor modalities, including challenging low-overlap conditions.
    \item We introduce a challenging cross-domain multi-view registration benchmark spanning five scene categories, on which our method substantially outperforms existing approaches in a zero-shot manner.
\end{itemize}

%%%%%%%%%%%%%%%%%%%%%%%%%%%%%%%%%%%%%%%%%%%%%%%%%%%%%%%%%%%%%%%%%%%%%%%%%%%%%%%%
\section{Related Work}
\label{sec:related}

\tightparagraph{Pairwise point cloud registration}has long relied on local feature matching with robust transformation estimators~\cite{fischler1981cacm,yang2020tro,lim2025icra-kissmatcher}.
Early approaches rely on hand-crafted local descriptors~\cite{rusu2009icra,tombari2010eccv-shot,salti2014cviu,dong2017jprs-bsc, he2016iros, mellado2014cgf-super4pcs} to establish correspondences. 
Modern methods~\cite{choy2019iccv, choy2020cvpr-dgr, gojcic2019cvpr-perfectmatch, bai2020cvpr-d3feat, huang2021cvpr-predator,qin2022cvpr-geotransformer,ao2021cvpr-spinnet, wang2022mm-yoho, wang2023pami-roreg, liu2024cvpr-eyoc, seo2025iccv, yao2024iccv-parenet, wiesmann2022ral-iros, yu2021neurips-cofinet, yu2023cvpr-roitr, li2022cvpr-lepard, shi2021ral, deng2019cvpr,bai2021cvpr-pointdsc, poiesi2023pami-gedi, yu2023cvpr-peal} replace or augment these with learned local features. 
Beyond explicit matches, correspondence-free and end-to-end approaches~\cite{aoki2019cvpr-pointnetlk,wang2019iccv-dcp,yew2020cvpr,wang2019neurips-prnet,bernreiter2021ral-phaser} directly supervise the transformation with differentiable assignment or iterative refinement. 
Our approach departs from both families: rather than seeking correspondences or iteratively refining poses, we learn a conditional velocity field that transports noise to the merged scene, from which rigid transformations are recovered.
Closer to our approach, DeepVCP~\cite{lu2019iccv} and DeepPRO~\cite{lee2021iccv-deeppro} generate virtual corresponding points near the target, but remain limited to pairwise registration with transformation initial guess or object-scale scenes.

\tightparagraph{Multi-view point cloud registration}is typically handled by first estimating local pairwise relative poses and then synchronizing the transformations via pose graph optimization (PGO)~\cite{grisetti2010titsmag, dellaert2012git}. In practice, optimization is performed using a factor graph with one node per scan and edges that encode relative pose measurements and their uncertainties under various robust objectives~\cite{theiler2015jprs,choi2015cvpr,dong2020jprs-whutls,gojcic2020cvpr,wang2023cvpr-sghr,jin2024cvpr-multiwaymosaicking, dong2018jprs}. 
By contrast, our single-stage approach aligns an arbitrary number of scans at once, enforcing multi-view consistency by construction. As a result, our method dispenses with the need for a separate PGO stage and avoids the quadratic costs from pairwise transformation estimation.
Nonetheless, our predictions can still serve as strong and time-efficient initializations for downstream PGO solvers that incorporate additional signals (e.g., from gravity, IMU, or GNSS) or task-specific constraints.

\tightparagraph{Generative modeling approaches for 3D data}leverage diffusion- and flow-based models to generate geometric structures~\cite{nunes2024cvpr, sun2025neurips, sanghi2023cvpr,wang2025iclr-puzzlefusionplusplus,xu2023cvpr-dzts,zeng2024cvpr-ppaw,bian2025iclr,guo2025neurips,zhang2025iccv,du2025cvpr,li2025iccv-garf,ren2024cvpr,jiang2023neurips-se3diffusion}. Both model generation as stochastic transport from source to target distributions: diffusion via iterative denoising, while flow-matching predicts velocity fields to iteratively transform data. These models have been applied to text-to-shape generation~\cite{sanghi2023cvpr,xu2023cvpr-dzts,zeng2024cvpr-ppaw}, 3D scene completion~\cite{nunes2024cvpr,zhang2025iccv,du2025cvpr}, and annotated data generation~\cite{ren2024cvpr,nunes2026pami,bian2025iclr,guo2025neurips}.

% Generative for pose estimation
More recent works leverage diffusion and flow-matching models to achieve point cloud registration~\cite{jiang2023neurips-se3diffusion,an2024heritagescience,sun2025neurips} as a way to overcome limitations of standard approaches. In DiffusionReg~\cite{jiang2023neurips-se3diffusion}, point cloud registration is formulated as a diffusion process on the SE(3) manifold, generating the corresponding rigid-body transformation between the source and target point clouds. 
Closest to our approach, RPF~\cite{sun2025neurips} applies Euclidean-space conditional flow matching to pairwise registration and multi-part shape assembly, but is limited to object-scale scenes (e.g., furniture and tableware).
RAP extends Euclidean-space flow matching to large-scale, cross-domain multi-view registration and achieves superior performance by scaling the training data, performing canonicalized generation conditioned on local embeddings, and rigidifying the flow sampling at test time.

%%%%%%%%%%%%%%%%%%%%%%%%%%%%%%%%%%%%%%%%%%%%%%%%%%%%%%%%%%%%%%%%%%%%%%%%%%%%%%%%
\section{Generative Point Cloud Registration}
\label{sec:main}

\begin{figure}[t]
    \centering
    \includegraphics[width=\linewidth]{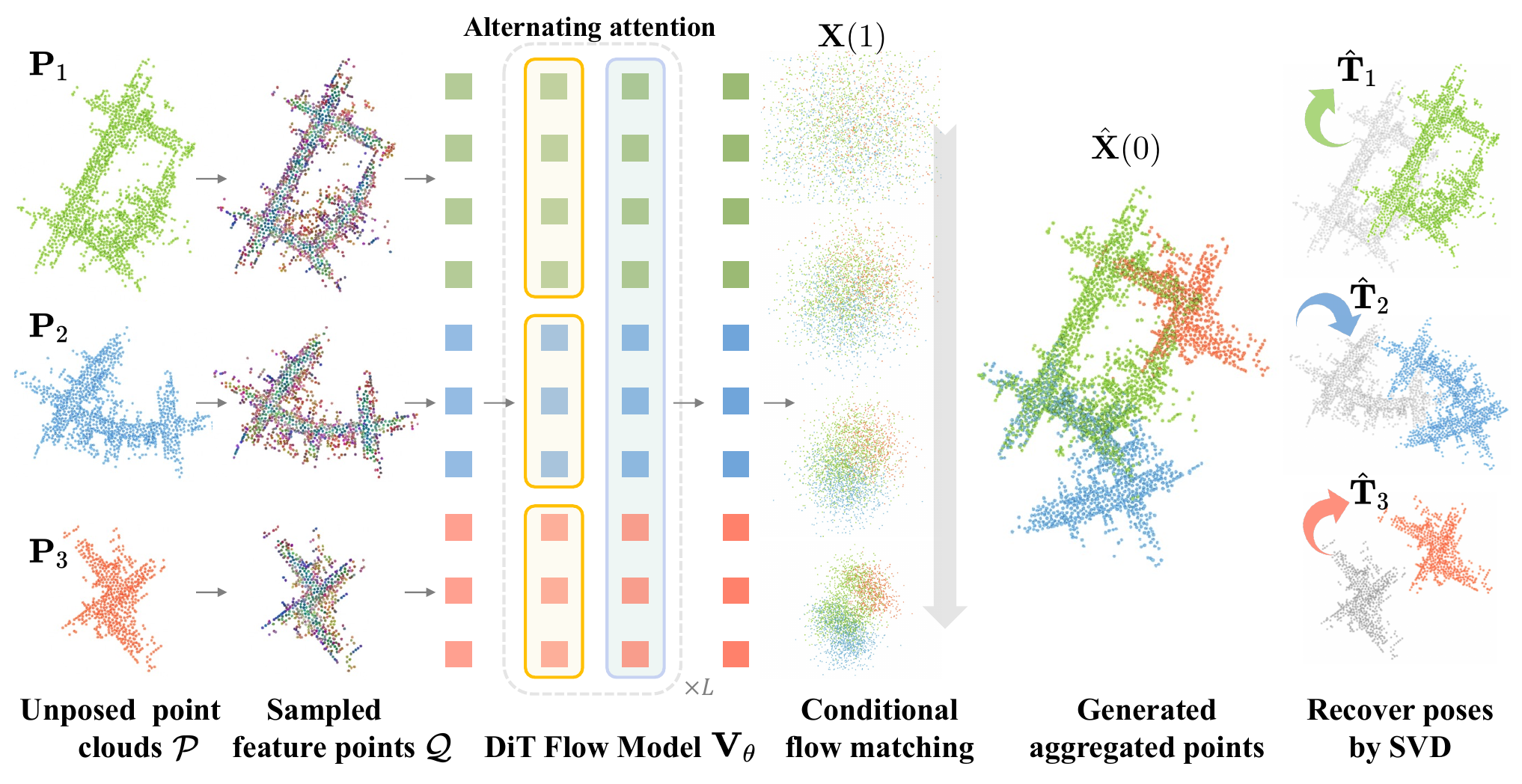}
    \vspace{-8pt}
    \caption{Overview of our approach to multi-view point cloud registration. Starting with unposed point clouds $\mathcal{P}$, we sample points $\mathcal{Q}$ with corresponding local features $\mathcal{F}$. We use a diffusion transformer (DiT) with alternating-attention blocks for conditional flow matching that generates the aggregated point cloud $\hat{\mathbf{X}}(0)$ from Gaussian noise $\mathbf{X}(1)$. Finally, we recover the individual transformations $\hat{\mathbf{T}}_i$ using singular value decomposition (SVD) from the aggregated point cloud and apply them to the original unposed point clouds to get the registered point clouds $\hat{\mathbf{P}}^r$. 
    The example illustrates submap registration on the KITTI~\cite{geiger2013ijrr} dataset.}
    \label{fig:model_overview}
    \vspace{-10pt}
\end{figure}

In this section, we present our generative approach to multi-view point cloud registration. We formulate registration as conditional flow matching, where a transformer-based model learns to directly generate the aggregated registered point cloud from unposed inputs.
We then describe our model architecture and training procedure, and how we enforce generation rigidity during inference.

\subsection{Problem Definition}
\label{subsec:problem_definition}

We consider the general multi-view point cloud registration problem. The input is a set of $N\geq 2$ unordered point clouds $\mathcal{P} = \{\mathbf{P}_i \in \RR^{3 \times M_i}\}_{i=1}^N$, where $M_i$ is the point count of the $i$-th point cloud.
These point clouds may come from individual LiDAR or depth-camera scans, or from accumulated point cloud maps built by SLAM or photogrammetry systems.
We do \emph{not} assume any initial guess of the transformation from the coordinate frame of each point cloud to a global frame, but we assume that the point clouds are observations of the same scene and can be registered into a single connected point cloud, even if the overlap is small.

The goal is to estimate registered point clouds $\mathcal{P}^r = \{\mathbf{P}_i^r \in \RR^{3 \times M_i}\}_{i=1}^N$ in a common global coordinate frame. 
In practice, there may be non-rigid deformations caused by effects such as motion distortion, dynamic objects, or SLAM drift. 
When these non-rigid effects are negligible, the registered point clouds can be transformed from the inputs by a set of rigid body transformations $\mathcal{T} = \{\mathbf{T}_i \in \mathrm{SE}(3)\}_{i=1}^{N}$. %which can be recovered via the Kabsch algorithm using SVD between $\mathcal{P}$ and $\mathcal{P}^r$.

\subsection{Flow Matching for Multi-view Registration}
\label{subsec:rpf}

Following RPF~\cite{sun2025neurips}, we formulate multi-view registration as a conditional generation problem and directly generate the registered point cloud $\mathcal{P}^r$ given the unposed input $\mathcal{P}$. The rigid transformations $\mathcal{T}$ are then recovered as a by-product.
We apply flow matching~\cite{liu2023iclr} directly to the 3D Euclidean coordinates of point clouds. The model learns to transport a 3D noised point cloud $\mathbf{X}(1) \in \RR^{3 \times M}$ sampled from a Gaussian $\mathcal{N}(\mathbf{0}, \mathbf{I})$ to a target point cloud $\mathbf{X}(0) \in \RR^{3 \times M}$ by learning a time-dependent velocity field $\nabla_t \mathbf{X}(t)$, parameterized by a neural network $\mathbf{V}_{\theta}(t, \mathbf{X}(t) \mid \mathbf{C})$ conditioned on $\mathbf{C}$, which will be detailed in Sec.~\ref{subsec:scaling}.
The forward process is a linear interpolation in 3D space between noise and the target, \textit{i.e}.,
\begin{equation}
    \mathbf{X}(t) = (1-t)\mathbf{X}(0) + t \mathbf{X}(1), \quad t \in [0, 1].
    \label{eq:fm}
\end{equation}
The flow model $\mathbf{V}_{\theta}$ is trained using the conditional flow matching loss~\cite{lipman2023iclr}.
For our registration task, the target $\mathbf{X}(0)$ is the aggregated registered point cloud
$\mathbf{P}^r = \bigcup_{i=1}^N \mathbf{P}_i^r$.

At inference time, we reconstruct the registered point cloud by numerically integrating the predicted velocity field $\mathbf{V}_\theta( t, \mathbf{X}(t) \mid \mathbf{C})$ from $t=1$ to $t=0$. In practice, we use $\kappa=10$ uniform Euler steps:
\begin{equation}
    \hat{\mathbf{X}}(t - \Delta t) = \hat{\mathbf{X}}(t) - \mathbf{V}_\theta(t, \hat{\mathbf{X}}(t) \mid \mathbf{C}) \Delta t,
    \label{eq:euler}
\end{equation}
with $\Delta t = 1/\kappa$. After integration, the resulting $\hat{\mathbf{X}}(0)$ approximates the registered point cloud.
We then partition $\hat{\mathbf{X}}(0)$ into per-view subsets and estimate the corresponding poses $\hat{\mathbf{T}}_i$ via the Kabsch algorithm~\cite{arun1987pami} using SVD.

\subsection{Rigidity-forcing Inference}
\label{subsec:rigidity}

The flow model alone does not guarantee per-view rigidity. While allowing non-rigid point motions increases the model's expressiveness, it can also drive sampling trajectories away from the training distribution defined in \eqref{eq:fm}. We therefore exploit the Euclidean nature of our flow formulation and introduce a \emph{rigidity-forcing} Euler integration that projects intermediate predictions onto per-view $\mathrm{SE}(3)$ orbits at each step. In addition, we empirically find that the resulting rigidity error during this procedure provides an effective criterion for selecting among multiple generations.

To illustrate Euler integration with rigidity-forcing, we define the per-view projection operator $\Pi$ of an estimate $\hat{\mathbf{X}}_i(0)$ onto the rigid orbit of the input $\mathbf{P}_i$, as
\begin{equation}
\Pi_{{\mathbf{P}}_i}\left(\hat{\mathbf{X}}_i(0)\right) :=\hat{\mathbf{R}}_i\,{\mathbf{P}}_i+\hat{\mathbf{t}}_i,
\label{eq:projected_clean}
\end{equation}
where $(\hat{\mathbf{R}}_i, \hat{\mathbf{t}}_i)$ is the optimal rigid transformation between $\mathbf{P}_i$ and $\hat{\mathbf{X}}_i(0)$, computed via the Kabsch algorithm~\cite{arun1987pami}.
Given the current state $\mathbf{X}(t)$ and the velocity $\mathbf{V}_\theta(t,\mathbf{X}(t)\!\mid\!\mathbf{C})$,
we extrapolate the registered point cloud estimate, as
\begin{equation}
\hat{\mathbf{Y}}(0) := \mathbf{X}(t)-t \,\mathbf{V}_\theta(t,\mathbf{X}(t)\!\mid\!\mathbf{C}).
\label{eq:x0hat}
\end{equation}
We now rigidify it to obtain the rigid projection of all views $\Pi_{{\mathcal{P}}}$ following \eqref{eq:projected_clean}
and compute the flow at the next step $t' \leftarrow t-\Delta t$ as
\begin{equation}
\mathbf{X}(t') := (1 - t') \,\Pi_{{\mathcal{P}}}\left(\hat{\mathbf{Y}}(0)\right) + t'\,\mathbf{X}(1).
\label{eq:rigid_step}
\end{equation}
We repeat the above sampling step until $t$ reaches $0$.

\subsection{Canonicalized Registration Pipeline}
\label{subsec:scaling}
Our method is designed to scale across scenes with diverse point densities, arbitrary coordinate frames, and a metric scale ranging from object-level scans to large outdoor environments. For this, we introduce a \textit{canonicalized keypoint-based} registration pipeline for our flow model.
The pipeline has four steps: (i) sampling a compact keypoint representation with local descriptors, (ii) canonicalizing all views into a shared similarity-invariant frame, (iii) conditioning a flow network on this representation to generate a canonical registered point cloud, and (iv) lifting the canonical prediction back to the original dense point clouds.

\tightparagraph{(i) Keypoint selection with local descriptors.}
Directly generating millions of points is computationally inefficient and unstable. Instead, we construct a compact, geometry-aware representation by sampling keypoints in each view and attaching local descriptors.
For each input point cloud $\mathbf{P}_i$, we first perform voxel downsampling with voxel size $v_d$, obtaining a reduced cloud $\mathbf{P}_i^{v}$. We then apply the farthest point sampling (FPS) to $\mathbf{P}_i^{v}$ to select $K_i$ keypoints,
forming $\mathbf{Q}_i \in \RR^{3 \times K_i}$, with the total keypoint count
$K := \sum_i K_i$ across all views. To obtain uniform coverage over the scene, we choose $K_i$ in proportion to the metric scale of $\mathbf{P}_i$.

To encode local geometry, for each sampled point in $\mathbf{Q}_i$ we define a local patch by a ball query of radius $r_s = 20 v_d$ in the reduced point cloud $\mathbf{P}_i^{v}$ and extract a local descriptor from the normalized points within this patch. We use the lightweight, rotation-invariant MiniSpinNet~\cite{ao2023cvpr-buffer, seo2025iccv} pretrained on 3DMatch~\cite{zeng2017cvpr}, yielding
$\mathbf{F}_i \in \RR^{32 \times K_i}$.
Concatenating all views, we obtain sampled points
$\mathcal{Q} = \{\mathbf{Q}_i \in \RR^{3 \times K_i}\}_{i=1}^N$ and their local features
$\mathcal{F} = \{\mathbf{F}_i \in \RR^{D \times K_i}\}_{i=1}^N$, which serve as a compressed yet informative representation of the dense input clouds $\mathcal{P}$.

\tightparagraph{(ii) Canonicalization of inputs and targets.}
To make training invariant to global pose and metric scale, we canonicalize both the conditioning representation and the flow target in a shared frame.
For each view $i$, we first translate the sampled points $\mathbf{Q}_i$ so that its center of mass is at the origin. We then compute a global scale factor $s$ as the longest edge length of the bounding box of the view with the most points, and scale all centered point sets by $1/s$, so that the entire scene fits in a unit cube. Finally, we apply a random 3D rotation to each centered, scaled cloud. This yields normalized unposed keypoints
$\bar{\mathcal{Q}} = \{\bar{\mathbf{Q}}_i \in \RR^{3 \times K_i}\}_{i=1}^N$ and corresponding similarity transforms
$\bar{\mathbf{T}}_i \in \mathrm{SIM}(3)$ that map the original $\mathbf{Q}_i$ to $\bar{\mathbf{Q}}_i$ in the canonical frame.

The training target is defined in the same canonical frame. We first transform the keypoints by the ground-truth poses
$\mathcal{\hat{T}}$ to obtain registered keypoints
$\mathcal{Q}^r = \{\mathbf{Q}_i^r\}_{i=1}^N$ and merge them into
$\mathbf{Q}^r = \bigcup_{i=1}^N \mathbf{Q}_i^r$.
We then (i) recenter $\mathbf{Q}^r$ at the origin, (ii) apply the same random rotation used for the view with the most points to fix a reference orientation, and (iii) scale by the global factor $s$, resulting in the normalized registered point cloud $\bar{\mathbf{Q}}^r$, which is set as the target $\mathbf{X}(0)$.

\tightparagraph{(iii) Conditional flow model.}
We adopt the Diffusion Transformer~\cite{peebles2023iccv} for $\mathbf{V}_{\theta}$, and, following VGGT~\cite{wang2025cvpr-vggt}, employ a transformer with alternating attention blocks (Fig.~\ref{fig:model_overview}). Specifically, we alternate per-view self-attention within each point cloud, which consolidates view-specific structure, with global attention over all point tokens to fuse information across views. Our model comprises $L=10$ alternating attention blocks with hidden dimension $d=512$ and $h=8$ attention heads, totaling 73 million parameters.

The model's condition $\mathbf{C} = f_\text{emb}(\bar{\mathcal{Q}}, \mathcal{F})$ is obtained via a linear feature embedder $f_\text{emb}$ with the concatenated local descriptors $\mathcal{F}$ and the positional encodings of the normalized coordinates $\bar{\mathcal{Q}}$.
The flow network $\mathbf{V}_{\theta}$ takes $\mathbf{X}(t)$ and $\mathbf{C}$ as input and is trained with the conditional flow matching loss~\cite{lipman2023iclr}.
Importantly, we do not condition the model on view indices, making the architecture view-count-agnostic and allowing it to generalize to larger numbers of views at test time than seen during training.

\tightparagraph{(iv) Lifting to dense registered point clouds.}
At inference, the model generates a canonical registered keypoint cloud $\hat{\mathbf{X}}(0)$.
Using the rigidity-forcing procedure in Sec.~\ref{subsec:rigidity}, we recover per-view rigid transformations $\hat{\mathbf{T}}_i$ that align $\bar{\mathbf{Q}}_i$ to the corresponding subsets of $\hat{\mathbf{X}}(0)$.
The overall transformation from the original input frame of view $i$ to the final registered frame is $\mathbf{T}_i = \hat{\mathbf{T}}_i \bar{\mathbf{T}}_i$,
which we apply to all points in the dense cloud $\mathbf{P}_i$.
Finally, we undo the global scaling by $s$ to obtain the registered point clouds at the metric scale.

%%%%%%%%%%%%%%%%%%%%%%%%%%%%%%%%%%%%%%%%%%%%%%%%%%%%%%%%%%%%%%%%%%%%%%%%%%%%%%%%
\section{Experimental Evaluation}
\label{sec:exp}

We present our experiments on pairwise and multi-view point cloud registration to show the capabilities of our proposed method named RAP.

\subsection{Experimental Setup}

\tightparagraph{Implementation details.}
We train our model for three days with about 120k iterations using 32 NVIDIA A100 GPUs with 80\,GB VRAM each.
We use Muon~\cite{liu2025arxiv-muon} as the optimizer with an initial learning rate of $2  \cdot 10^{-3}$ for matrix-like parameters and $2 \cdot 10^{-4}$ for vector-like parameters.

Unlike VGGT~\cite{wang2025cvpr-vggt} or RPF~\cite{sun2025neurips}, our training samples have varying numbers of views $N$ and feature point (token) counts~$K$. To efficiently train under this irregular data setup, we devise a dynamic batching strategy that allocates samples to each GPU with a suitable batch size. We set the maximum token count per batch on one GPU to 110,000.

\tightparagraph{Training data.}
Our training requires only a set of point clouds under the same reference frame without any annotations for keypoints or correspondences. This makes it straightforward to scale up training as any dataset providing point clouds and accurate sensor poses can be used.

For our model, we curate over 100k multi-view registration instances from 17 diverse datasets spanning outdoor, indoor, and object-centric settings.
We train on 12 outdoor LiDAR datasets KITTI~\cite{geiger2012cvpr}, KITTI360~\cite{liao2022pami}, Apollo~\cite{huang2018cvprws}, nuScenes~\cite{caesar2020cvpr}, MulRAN~\cite{kim2020icra}, Boreas~\cite{burnett2023ijrr-boreas}, Oxford Spires~\cite{tao2025ijrr-oxfordspires}, VBR~\cite{brizi2024icra-vbr}, UrbanNav~\cite{hsu2023navi-urbannav}, WildPlace~\cite{knights2023icra-wildplaces}, HeLiPR~\cite{jung2024ijrr}, and KITTI-Carla~\cite{deschaud2021arxiv-kitticarla}.
We also use 4 indoor depth camera datasets 3DMatch~\cite{zeng2017cvpr}, NSS~\cite{sun2025jprs}, ScanNet~\cite{dai2017cvpr}, and ScanNet++~\cite{yeshwanth2023iccv-scannetpp}, and the object-centric dataset ModelNet~\cite{wu2015cvpr}.
Each sample consists of $N$ views of point clouds, where $2 \leq N \leq 16$.
We split the samples into training and validation sets with an approximate ratio of 9:1 while we also exclude sequences used for testing in common registration benchmarks from the training set.
The datasets span diverse scenes across multiple continents, captured by LiDAR and depth cameras with varying resolutions and fields of view.
We curate both single-frame and sequence-accumulated submap samples. We require all point clouds in a sample to form a connected overlap graph, but deliberately include hard samples with very low pairwise overlap to improve the model's robustness to low-overlap scenarios.

\subsection{Pairwise Registration Evaluation}

We first evaluate our model on pairwise registration across six widely used benchmarks: ModelNet~\cite{wu2015cvpr} for object-centric scenes; 3DMatch~\cite{zeng2017cvpr}, 3DLoMatch~\cite{huang2021cvpr-predator}, and NSS~\cite{sun2025jprs} for indoor depth-camera scenarios; and ETH~\cite{pomerleau2012ijrr} and KITTI~\cite{geiger2012cvpr} for outdoor LiDAR scenarios.
Following prior work, we evaluate performance using registration success rate~(\%), computed with thresholds on correspondence RMSE for 3DMatch and 3DLoMatch, and on translation and rotation error for the remaining datasets. We adopt the success thresholds used in previous studies~\cite{zeng2017cvpr, sun2025jprs, seo2025iccv}; details are provided in the supplementary material.
We compare against three conventional baselines based on hand-crafted features (FPFH\allowbreak+FGR~\cite{zhou2016eccv}, FPFH\allowbreak+TEASER~\cite{yang2020tro}, KISS-Matcher~\cite{lim2025icra-kissmatcher}) and six learning-based methods (FCGF~\cite{choy2019iccv}, Predator~\cite{huang2021cvpr-predator}, GeoTransformer~\cite{qin2022cvpr-geotransformer}, BUFFER~\cite{ao2023cvpr-buffer}, PARENet~\cite{yao2024iccv-parenet}, BUFFER-X~\cite{seo2025iccv}).

\cref{table:pairwise_success_rates} shows that our model ranks first on half of the benchmarks and achieves the best average performance across all six benchmarks compared to state-of-the-art methods. 
We note that our model is trained on more diverse data, while some learning-based baselines are trained solely on 3DMatch or KITTI. 
We provide additional results on pairwise registration under low overlap in the supplementary material.

\begin{table}[t!]
  \centering
  \caption{Pairwise registration success rate (\%) on six commonly used pairwise registration benchmarks for object-centric, indoor, and outdoor scenarios. The best result is in \textbf{bold}, and the second best is \underline{underscored}.}
  \vspace{-5pt}
  \resizebox{0.99\linewidth}{!}{%
  {\scriptsize
  % Set width per column (Method, ModelNet, 3DMatch, 3DLoMatch, NSS, ETH, KITTI, Avg):
  \def\cwMethod{10em}\def\cwModelNet{4.8em}\def\cwA{4.8em}\def\cwB{4.8em}\def\cwC{4.8em}\def\cwD{4.8em}\def\cwE{4.8em}\def\cwAvg{4.8em}
  \begin{tabular}{>{\raggedright\arraybackslash}p{\cwMethod}|>{\centering\arraybackslash}p{\cwModelNet}>{\centering\arraybackslash}p{\cwA}>{\centering\arraybackslash}p{\cwB}>{\centering\arraybackslash}p{\cwC}>{\centering\arraybackslash}p{\cwD}>{\centering\arraybackslash}p{\cwE}|>{\centering\arraybackslash}p{\cwAvg}}
    \toprule
    \midrule
    Method & ModelNet & 3DMatch & 3DLoMatch & NSS & ETH & KITTI & Avg. \\ \midrule
    FPFH+FGR~\cite{zhou2016eccv} & 84.04 & 62.53 & 15.42 & 30.82 & 91.87 & 98.74 & 63.90 \\
    FPFH+TEASER~\cite{yang2020tro} & 86.10 & 52.00 & 13.25 & 25.78 & 93.69 & 98.92 & 61.62 \\
    KISS-Matcher~\cite{lim2025icra-kissmatcher} & 87.12 & 67.22 & 20.44 & 53.69 & 96.77 & \textbf{100.0} & 70.87 \\
    \midrule
    FCGF~\cite{choy2019iccv} & 16.51 & 88.18 & 40.09 & 42.86 & 55.53 & 98.92 & 58.68 \\
    Predator~\cite{huang2021cvpr-predator} & 84.36 & 90.60 & 62.40 & \underline{92.99} & 54.42 & \underline{99.82} & 80.77 \\
    GeoTransformer~\cite{qin2022cvpr-geotransformer} & 86.26 & 92.00 & 75.00 & 55.59 & 71.53 & \underline{99.82} & 80.03 \\
    BUFFER~\cite{ao2023cvpr-buffer} & 92.42 & 92.90 & 71.80 & 72.44 & 99.30 & 99.64 & 88.12 \\
    PARENet~\cite{yao2024iccv-parenet} & 66.14 & 95.00 & \textbf{80.50} & 45.07 & 69.42 & \underline{99.82} & 75.99 \\
    BUFFER-X~\cite{seo2025iccv} & \textbf{99.84} & \underline{95.58} & 74.18 & 85.60 & \textbf{99.72} & \underline{99.82} & \underline{92.46} \\
    \midrule
    RAP (Ours) & \underline{99.13} & \textbf{95.90} & \underline{78.78} & \textbf{96.27} & \underline{99.44} & \textbf{100.0} & \textbf{94.92} \\
    \midrule
    \bottomrule
  \end{tabular}%
  }}
  \label{table:pairwise_success_rates}
  \vspace{-8pt}
\end{table}

\subsection{Multi-view Registration Evaluation}

\begin{figure}[t]
  \centering
  \includegraphics[width=0.97\linewidth]{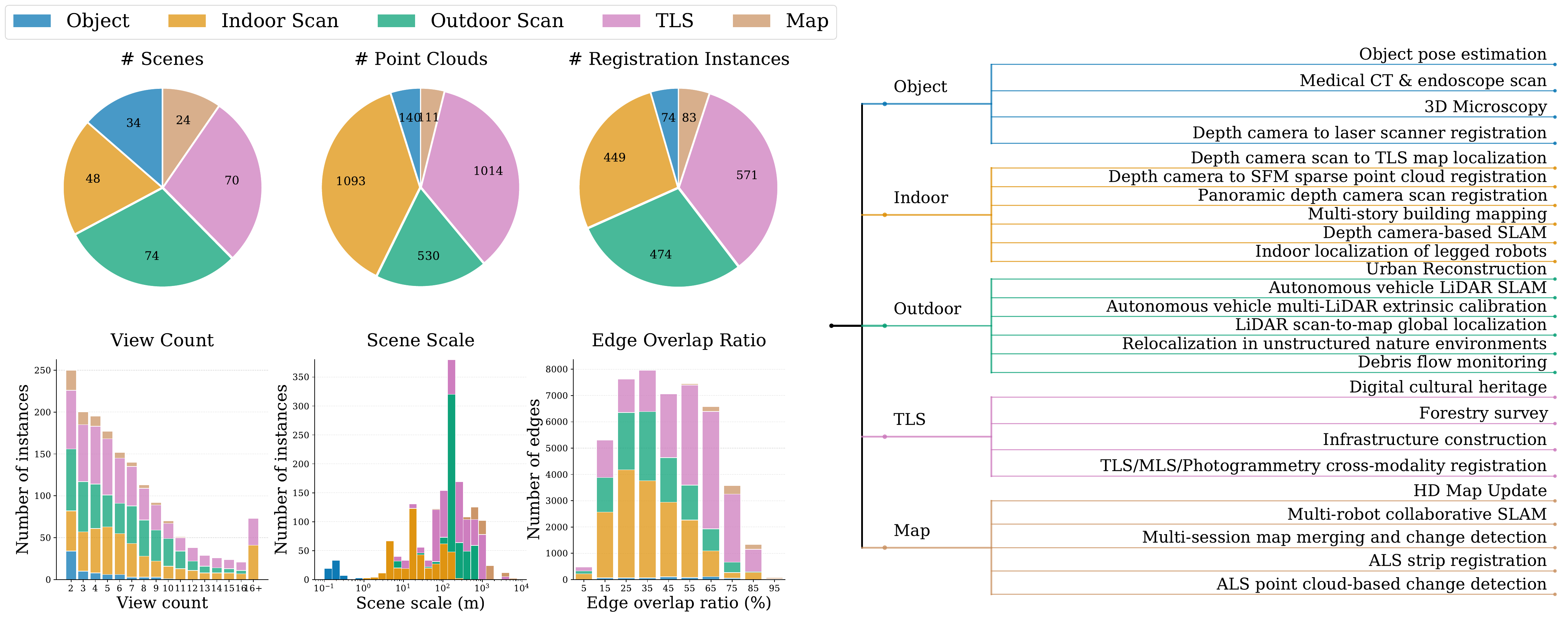}\\[-1pt]
  \includegraphics[width=0.97\linewidth]{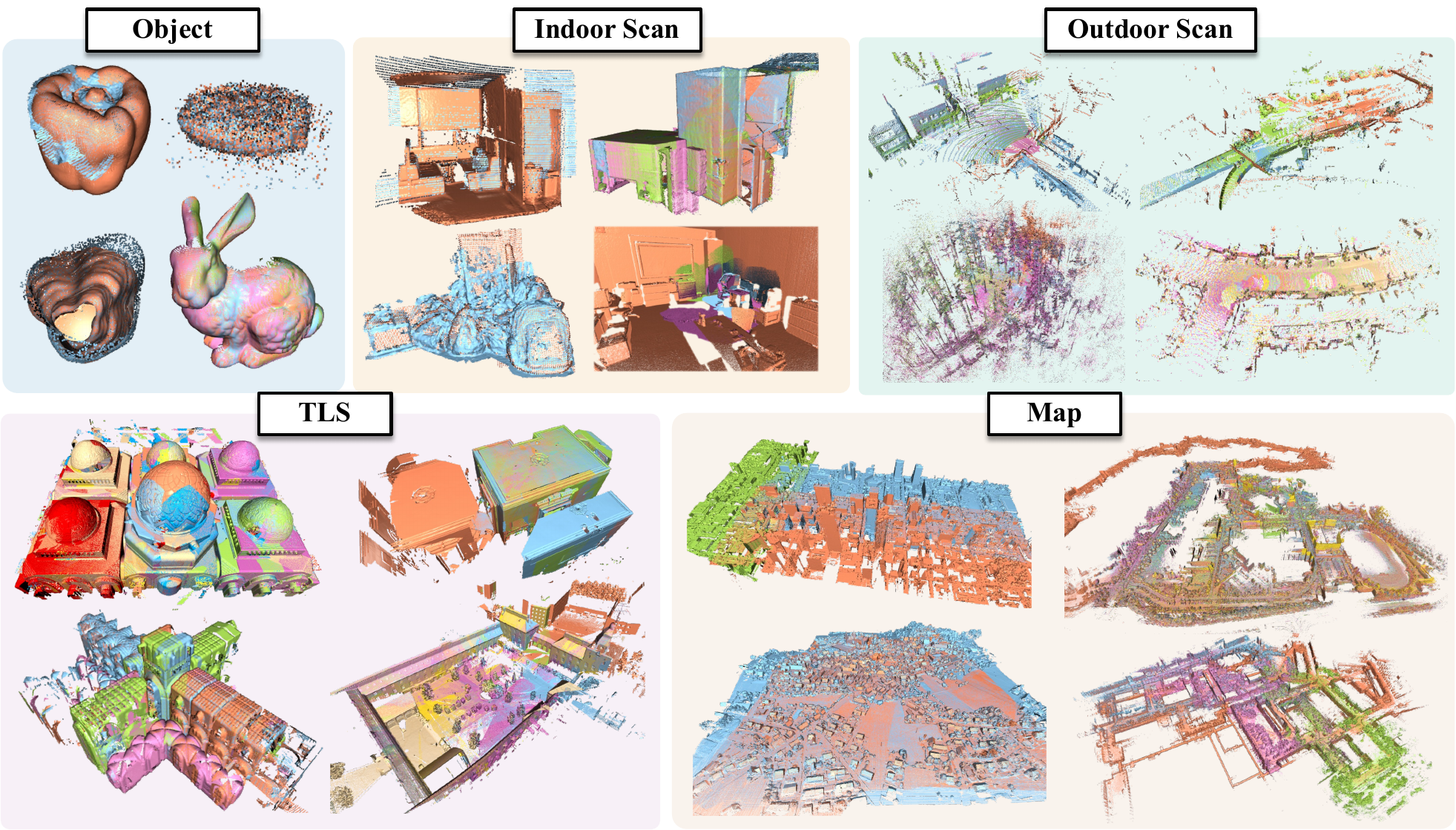}
  \caption{Overview of our proposed cross-domain multi-view registration benchmark, showing key dataset statistics (top-left), covered application domains (top-right), and representative samples from each scenario category (bottom). Different colors of the point cloud denote different views.}
  \label{fig:rapbench_info}
  \vspace{-10pt}
\end{figure}

\tightparagraph{Cross-domain multi-view registration benchmark.}
We identify a gap in the evaluation of point cloud registration methods on challenging real-world data. While pairwise registration success rates on established benchmarks such as ModelNet, ETH, and KITTI are nearly saturated (see \cref{table:pairwise_success_rates}), methods that perform well on these benchmarks often fail to generalize to practical applications. Furthermore, while some prior works use indoor depth-camera datasets~\cite{zeng2017cvpr, dai2017cvpr, sun2025jprs, choi2015cvpr} for multi-view registration evaluation, no cross-domain benchmark exists covering diverse scales, scenarios, and sensor modalities.

To address this gap, we propose a challenging cross-domain multi-view registration benchmark (\cref{fig:rapbench_info}). It aggregates point cloud datasets from diverse sources, most of which were not designed for registration evaluation, and spans five scene categories: object, indoor scan, outdoor scan, terrestrial laser scanning (TLS), and map. Crucially, none of these datasets overlap with our training mixture, making evaluation zero-shot. The benchmark covers practical applications including robot grasping, indoor relocalization, multi-session map merging, collaborative SLAM, TLS surveying, and remote sensing change detection. Dataset sources and curation details are provided in the supplementary material.

\tightparagraph{Evaluation metrics and baselines.}
We adopt the edge success rate and graph success rate as our main evaluation metrics. The edge success rate is the mean over all valid edges (i.e., pairs with non-zero overlap) in a sample, while the graph success rate requires every valid edge in the multi-view registration graph to be successful.
To handle diverse scene scales, we normalize translation error by the longest axis of the ground-truth bounding box; a registration succeeds if the normalized translation error is below 2.5\% and rotation error below 15\textdegree, a threshold reliably correctable by ICP~\cite{besl1992pami}. Stricter threshold results (0.5\%, 3\textdegree) are in the supplementary.

We select the top-3 performing conventional and learning-based pairwise registration methods from \cref{table:pairwise_success_rates} as baselines, each paired with a robust and efficient implementation of pose graph optimization~\cite{choi2015cvpr} to form a complete two-stage multi-view registration pipeline.
For a fair comparison, we retrain BUFFER-X, the best-performing baseline in the pairwise evaluation, on the same extended training data as RAP, denoting this variant BUFFER-X$^*$.
We also include two dedicated multi-view methods: SGHR~\cite{wang2023cvpr-sghr}, which uses learned overlap scores and history-reweighting iterative optimization, and RPF~\cite{sun2025neurips}, a flow-matching method targeting object-level shape assembly.

\tightparagraph{Experiment results.}
As shown in \cref{tab:rapbench_eval}, RAP achieves the best performance across all five scenario categories by a large margin, improving the overall edge success rate from 57.7\% to 85.9\% and graph success rate from 36.5\% to 77.3\% compared to the strongest pairwise baseline BUFFER-X$^*$.
Training BUFFER-X with more diverse data yields a modest gain over the original model, yet still falls significantly short of RAP, indicating that data scaling alone is insufficient without a more capable model architecture and training target.
Other learning-based methods (Predator~\cite{huang2021cvpr-predator}, BUFFER~\cite{ao2023cvpr-buffer}, SGHR~\cite{wang2023cvpr-sghr}, RPF~\cite{sun2025neurips}) fail to generalize across domains; RPF in particular struggles as it operates on raw points without local descriptors, which does not scale to large scenes.
KISS-Matcher~\cite{lim2025icra-kissmatcher} is relatively competitive (46.2\%/29.2\%) thanks to its scalable hand-crafted features and robust transformation estimator, but is still far behind RAP.
RAP w/o rigidity forcing (RF) already surpasses all baselines, with test-time rigidity forcing integration providing further consistent improvement.

Notably, RAP achieves the shortest runtime at 8.9\,s per sample, outpacing all baselines including KISS-Matcher+PGO (18.4\,s), BUFFER-X$^*$+PGO (36.0\,s), and SGHR (49.7\,s). This efficiency stems from RAP's single-stage design: unlike two-stage pipelines that first perform pairwise registration on all or a selected subset of likely-overlapping pairs (as in SGHR) and then run pose graph optimization, RAP directly generates the registered point cloud in one forward pass without any pairwise stage. We further analyze how runtime scales with view count and token count in \cref{fig:rapbench_runtime_compare}.

% Evaluation comparison on the cross-domain multi-view registration benchmark

\begin{table}[!t]
  \centering
  \caption{Comparison of the zero-shot testing performance for multi-view registration on the cross-domain multi-view registration benchmark. We report the registration success rate (\%) calculated for edges and graphs (with a threshold of 2.5\% for normalized translation error and 15$^\circ$ for rotation error) in different scenarios as well as the average runtime. The best result is in \textbf{bold}, and the second best is \underline{underscored}. }
  \label{tab:rapbench_eval}
  \vspace{-6pt}
  \resizebox{\linewidth}{!}{%
  {\scriptsize
  \begin{tabular}{l|c|cc|cc|cc|cc|cc|>{\columncolor{gray!10}}c>{\columncolor{gray!10}}c|c}
  \toprule
  \midrule
  \multicolumn{2}{c|}{\multirow{2}{*}{Method}} & \multicolumn{2}{c|}{Object} & \multicolumn{2}{c|}{Indoor} & \multicolumn{2}{c|}{Outdoor} & \multicolumn{2}{c|}{TLS} & \multicolumn{2}{c|}{Map} & \multicolumn{2}{c|}{\cellcolor{gray!10}All} & \multirow{2}{*}{Runtime (s)} \\
  \cmidrule(lr){3-4} \cmidrule(lr){5-6} \cmidrule(lr){7-8} \cmidrule(lr){9-10} \cmidrule(lr){11-12} \cmidrule(lr){13-14}
  \multicolumn{2}{c|}{} & edge & graph & edge & graph & edge & graph & edge & graph & edge & graph & \cellcolor{gray!10}edge & \cellcolor{gray!10}graph & \\
  \midrule
  FPFH+FGR~\cite{zhou2016eccv} & \multirow{7}{*}{\rotatebox[origin=c]{270}{+PGO}} & 14.4 & 6.8 & 2.8 & 1.3 & 4.3 & 1.1 & 5.6 & 2.0 & 9.0 & 7.7 & 5.5 & 2.2 & 19.7  \\
  FPFH+TEASER~\cite{yang2020tro} &  & 18.3 & 13.5 & 8.7 & 5.3 & 13.9 & 9.6 & 17.3 & 12.4 & 30.2 & 19.4 & 16.7 & 11.4 & 22.2 \\
  KISS-Matcher~\cite{lim2025icra-kissmatcher} & & 38.1 & 29.7 & 30.5 & 14.0 & 33.4 & 19.4 & 63.9 & 42.5 & \underline{81.8} & \underline{70.3} & 46.2 & 29.2 & 18.4 \\
  \cmidrule(lr){1-1} \cmidrule(lr){3-15}
  Predator~\cite{huang2021cvpr-predator} & & 2.8 & 1.9 & 9.0 & 5.9 & 7.2 & 5.1 & 4.3 & 3.6 & 8.2 & 5.7 &  6.8 & 4.6 & 51.8 \\
  BUFFER~\cite{ao2023cvpr-buffer} & & 2.3 & 1.1 & 25.5 & 12.0 & 16.7 & 9.8 & 28.7 & 16.9 & 3.9 & 1.5 & 22.5 & 11.8 & 250.3 \\
  BUFFER-X~\cite{seo2025iccv} & & 67.6 & 56.8 & 43.4 & 22.3 & 44.6 & 31.0 & 63.9 & 42.1 & 26.9 & 17.2 & 50.4 & 32.4 & 36.0 \\
  BUFFER-X*~\cite{seo2025iccv} & & \underline{70.3} & \underline{59.2} & \underline{47.3} & \underline{25.1} & \underline{54.0} & \underline{37.1} & \underline{69.2} & \underline{46.8} & 33.6 & 21.5 & \underline{57.7} & \underline{36.5} & 36.0 \\
  \midrule
  \multicolumn{2}{l|}{SGHR~\cite{wang2023cvpr-sghr}} & 12.6 & 8.1 & 28.9 & 22.0 & 19.8 & 11.0 & 24.6 & 17.6 & 15.1 & 10.9 & 24.3 & 17.2 & 49.7 \\
  \multicolumn{2}{l|}{RPF~\cite{sun2025neurips}} & 2.5 & 1.7 & 0.8 & 0.5 & 0.7 & 0.3 & 0.9 & 0.3 & 1.3 & 0.9 & 1.0 & 0.6 & \underline{11.5} \\
  \midrule
  \multicolumn{2}{l|}{RAP (Ours)}        & \textbf{86.3} & \textbf{85.1} & \textbf{75.1} & \textbf{59.6} & \textbf{86.6} & \textbf{80.6} & \textbf{92.2} & \textbf{84.6} & \textbf{94.0} & \textbf{91.6} & \textbf{85.9} & \textbf{77.3} & \textbf{8.9} \\
  \multicolumn{2}{l|}{RAP (Ours) w/o RF} & 83.1 & 82.4 & 72.3 & 54.7 & 84.7 & 76.6 & 90.1 & 80.0 & 89.7 & 86.8 & 83.6 & 72.9 & 8.9 \\
  \midrule
  \bottomrule
  \end{tabular}%
  }}
  \vspace{-4pt}
\end{table}

\begin{figure}[!t]
  \centering
  \includegraphics[width=1.0\linewidth]{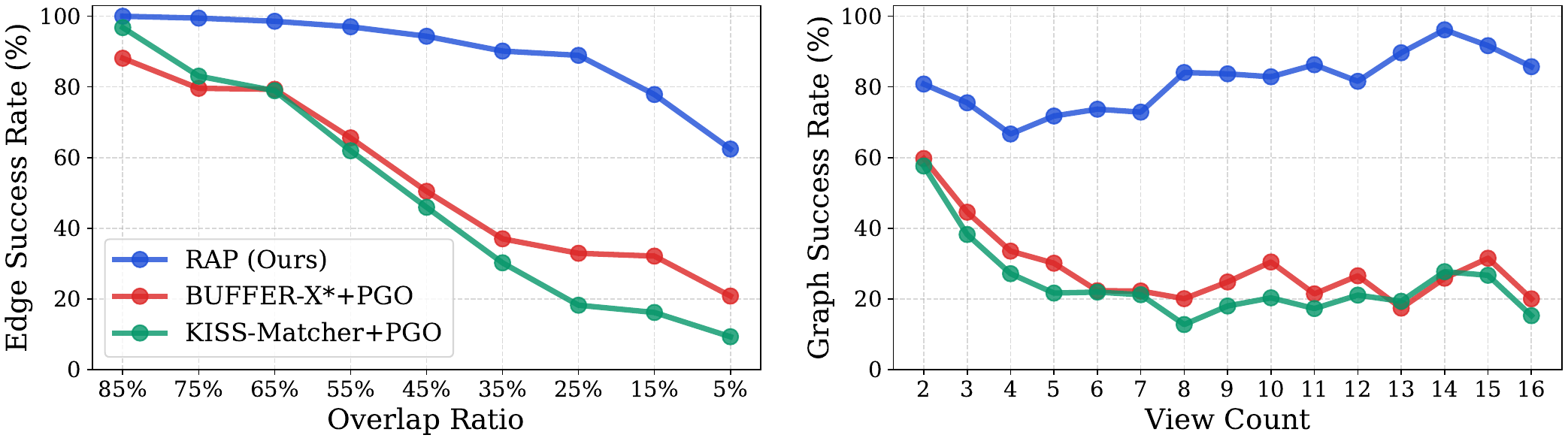}
  \vspace{-10pt}
  \caption{Multi-view registration performance analysis on the cross-domain multi-view registration benchmark. Left: Edge success rate (mean over pairs) vs.  edge overlap ratio. Right: Graph success rate (all pairs must succeed) vs. number of views. RAP outperforms BUFFER-X* and KISS-Matcher across overlap and view counts.}
  \label{fig:rapbench_sr_compare}
  \vspace{-6pt}
\end{figure}

\cref{fig:rapbench_sr_compare} provides an in-depth analysis of RAP's performance as a function of overlap ratio and view count, comparing against the two best-performing baselines.
As shown on the left, RAP maintains a consistently higher edge success rate than BUFFER-X$^*$+PGO and KISS-Matcher+PGO across all overlap ratios. The performance gap is modest at high overlap but widens substantially as overlap decreases. At an overlap ratio of 20\%, RAP still achieves above 80\% success rate, while the baselines drop below 40\%, demonstrating the superior robustness of our method under low-overlap conditions.
The right plot shows that the graph success rate of RAP remains stable and even improves with more views, while the baselines deteriorate more steeply.
This highlights a key weakness of two-stage pipelines: erroneous pairwise registrations propagate as outliers into the pose graph, which pose graph optimization cannot reliably resolve. In contrast, our method does not adjust the  per-view transformations explicitly. Instead, it performs an implicit global adjustment at the point level, transporting all keypoints across all views simultaneously toward a globally consistent configuration. Additional views therefore provide complementary geometric evidence rather than additional sources of pairwise error.

We provide additional multi-view registration results on 3DMatch~\cite{zeng2017cvpr} and ScanNet~\cite{dai2017cvpr} in the supplementary material.

\subsection{Ablation Studies}

We evaluate our model under different inference and architectural configurations on 3DMatch and 3DLoMatch for pairwise registration, and the cross-domain multi-view registration benchmark for multi-view registration. The results are summarized in Table~\ref{tab:ablation}.

Under the default setting [A] with test-time rigidity forcing and $\kappa=10$ step Euler integration, our model achieves the highest success rates across all datasets. Setting [B] removes rigidity forcing during sampling, and setting [C] reduces the number of flow integration steps $\kappa$ from 10 to 1. Both degrade performance, with [B] showing that enforcing rigidity during sampling is crucial and [C] confirming a clear accuracy--speed trade-off with fewer steps. Note that even with a one-step generation, our model outperforms all the baseline methods on the benchmark. 

Settings [D]--[F] study the impact of model design. Setting [D] uses only the sampled keypoint coordinates as conditions without local descriptors, and its performance drops sharply compared to [A], especially on the benchmark. Highlighting the importance of local geometric features. Settings [E] and [F] evaluate smaller transformer models with $L$ decreasing from 10 to 8 and 6, respectively. As expected, reducing model capacity induces a clear performance drop.
Additional ablations and results are provided in the supplementary material.

\begin{table}[t]
\centering
\caption{Ablation studies. We report the registration success rate (\%) on datasets for both pairwise and multi-view registration tasks. Best results are shown in \textbf{bold}. On the right figure, we compare the edge success rate of [A]-[C] on the cross-domain multi-view registration benchmark for five different scenario categories.}
\vspace{-8pt}
\begin{minipage}[c]{0.64\linewidth}
  \centering
  \resizebox{\linewidth}{!}{%
  {\footnotesize
  \begin{tabular}{l|cc|cc}
  \toprule \midrule
  \multirow{2}{*}{\textbf{Setting}} & \multicolumn{2}{c|}{Pairwise} & \multicolumn{2}{c}{Multi-view} \\
  \cmidrule(lr){2-3} \cmidrule(lr){4-5}
  & 3DMatch & 3DLoMatch & Edge & Graph \\ \midrule
  {[A]} Ours  & \textbf{95.90} & \textbf{78.78} & \textbf{85.92} & \textbf{77.31} \\
  \midrule
  \textbf{Inference} & & & & \\
  {[B]} w/o rigidity forcing (RF) & 94.65 & 74.63 & 83.57 & 72.88 \\
  {[C]} w/ 1-step generation & 90.35 & 61.05 & 72.87 & 50.32 \\
  \midrule
  \textbf{Model} & & & & \\
  {[D]} w/o local feature extraction  & 83.28 & 50.05 & 43.07 & 26.85 \\
  {[E]} w/ $L=8$ transformer blocks \, & 93.24 & 71.63 & 81.08 & 72.19 \\
  {[F]} w/ $L=6$ transformer blocks \, & 90.90 & 57.59 & 65.64 & 54.35 \\
  \midrule \bottomrule
  \end{tabular}%
  }}
\end{minipage}%
\hfill
\begin{minipage}[c]{0.3\linewidth}
  \centering
  \includegraphics[width=\linewidth]{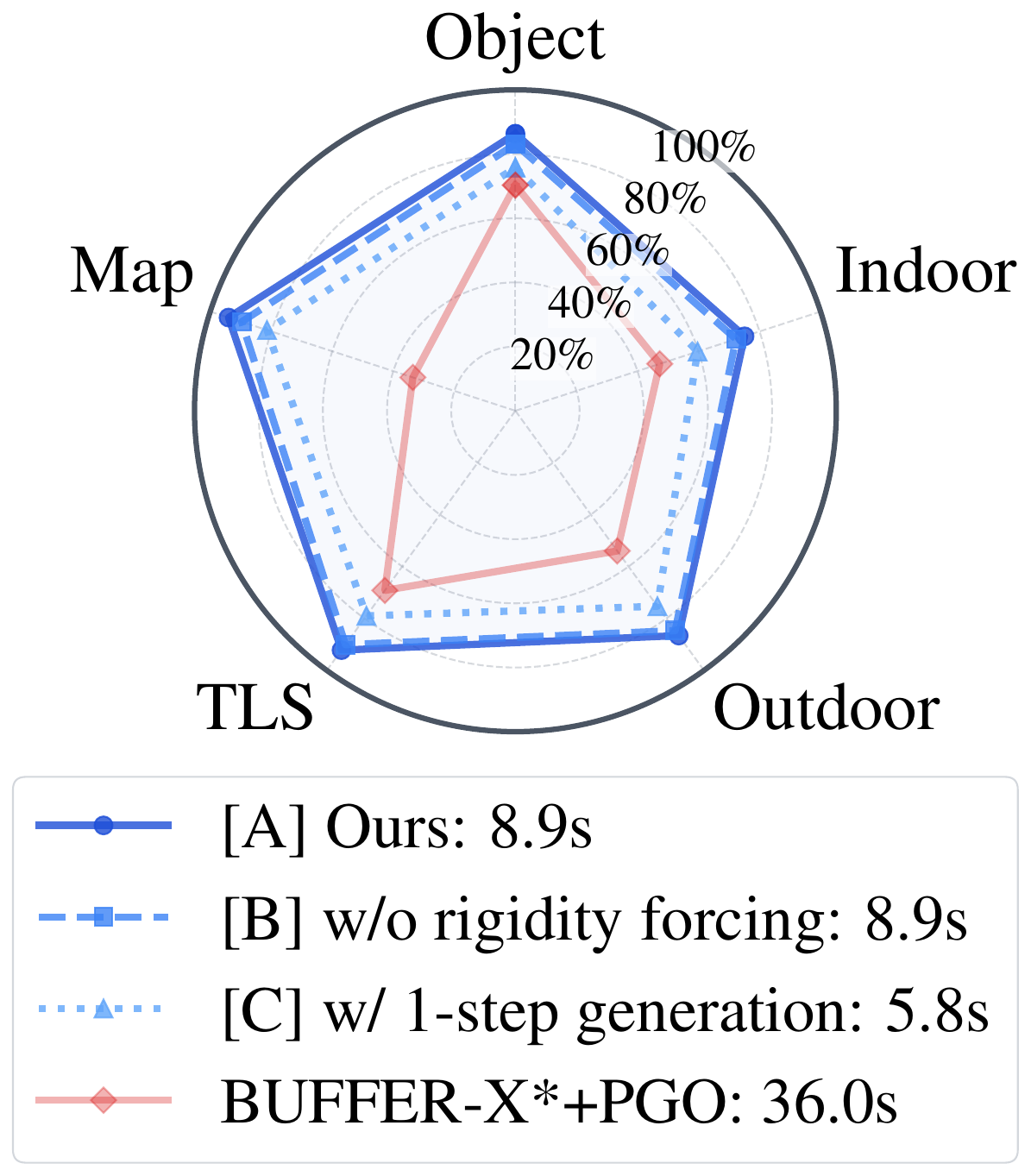}
\end{minipage}
\vspace{-10pt}
\label{tab:ablation}
\end{table}

\subsection{Runtime}

\begin{figure}[!t]
  \centering
  \includegraphics[width=1.0\linewidth]{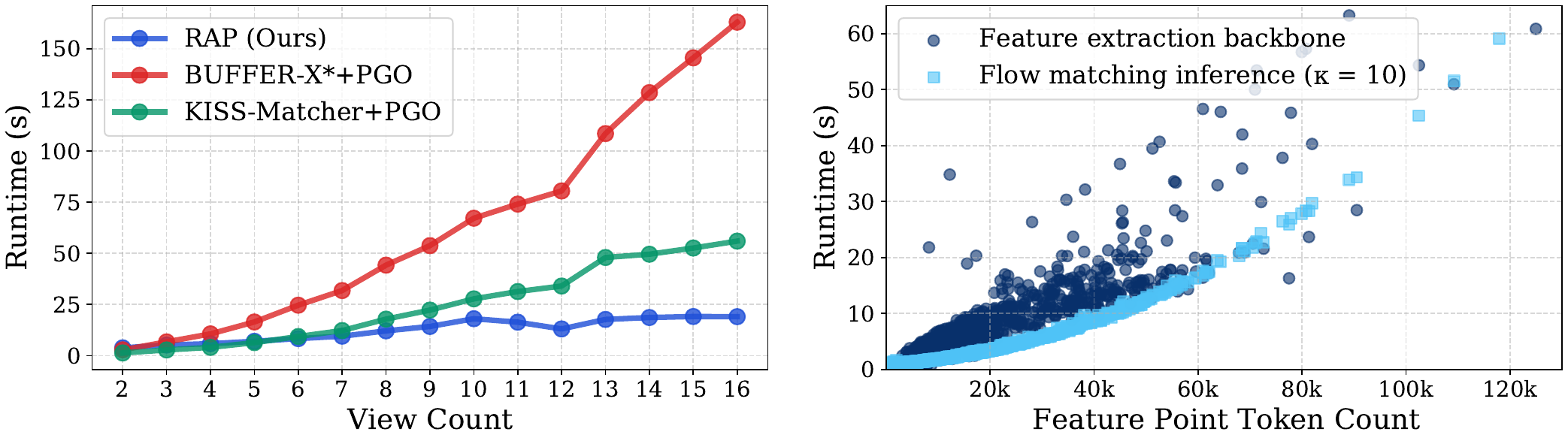}
  \vspace{-10pt}
  \caption{Multi-view registration run time analysis on the cross-domain multi-view registration benchmark. Left: Mean runtime per sample vs. view count for RAP, BUFFER-X*+PGO, and KISS-Matcher+PGO. Right: RAP runtime vs. total feature point token count fed into the transformer, split into feature extraction backbone and flow-matching inference time. Runtimes reported on a single Nvidia A5000 GPU excluding IO.}
  \label{fig:rapbench_runtime_compare}
  \vspace{-6pt}
\end{figure}

We further evaluate the inference time of our model on a local workstation with a single NVIDIA A5000 GPU of 24\,GB VRAM. 
As shown in \cref{fig:rapbench_runtime_compare} (left), at two views, RAP is slightly slower than KISS-Matcher and BUFFER-X$^*$ due to the flow-matching inference overhead. As the number of views grows, however, the baselines scale poorly due to the cost of all-pairs pairwise registration and pose graph optimization, while RAP's runtime remains nearly linear, quickly becoming the fastest option overall.
The right plot breaks down RAP's compute into two components: feature extraction and flow-matching inference. Feature extraction dominates at low token counts but scales linearly with the total number of feature point tokens, while flow-matching inference grows more steeply due to the quadratic complexity of global attention. Overall, the runtime scales gracefully, and the number of Euler steps $\kappa$ can provide a direct accuracy--speed trade-off at flow matching generation time.

%%%%%%%%%%%%%%%%%%%%%%%%%%%%%%%%%%%%%%%%%%%%%%%%%%%%%%%%%%%%%%%%%%%%%%%%%%%%%%%%
\section{Conclusion}
\label{sec:conclusion}

We presented RAP, a generative flow-matching model for single-stage multi-view point cloud registration. By casting registration as conditional point flow generation with an alternating-attention transformer, RAP achieves holistic multi-view reasoning without a pairwise stage or pose graph optimization, yielding strong robustness under low overlap.
Trained on over 100k multi-view registration instances from 17 diverse datasets and evaluated zero-shot on our proposed cross-domain benchmark, RAP achieves state-of-the-art performance across object, indoor, outdoor, TLS, and map domains. A rigidity-enforcing sampler provides further test-time gains. We hope RAP serves as a step toward a universal registration foundation model for applications such as SLAM, 3D reconstruction, and robotic manipulation.

\tightparagraph{Limitations and future work}
Our approach assumes that the scans to be registered are recorded in the same environment, like other feed-forward reconstruction methods~\cite{wang2025cvpr-vggt,wang2025arxiv-pi3}. By modeling flow in Euclidean space rather than the transformation group, our method could potentially handle non-rigid transformations, though this remains unexplored. Future work may extend to scene flow estimation~\cite{huang2022eccv} and merging point maps from photogrammetry~\cite{schoenberger2016cvpr-sfm} and feed-forward 3D reconstruction~\cite{wang2025cvpr-vggt,wang2025arxiv-pi3}.

% \section*{Acknowledgements}
% Please insert your acknowledgments here.

% ---- Supplementary Material (Appendix) ----
\newpage
\appendix
\section*{Supplementary Material}
\addcontentsline{toc}{section}{Supplementary Material}

\setcounter{section}{0}
\renewcommand{\thesection}{\Alph{section}}

\section{Overview}
\label{sec:supp_overview}

In the supplementary material, we provide the following:

\begin{itemize}
    \item Implementation details including point cloud sampling and feature extraction, flow model architecture and training, training data curation, pairwise registration testing data, and evaluation metrics (Section~\ref{sec:implementation_details})
    \item Details on the selected baseline methods used in our experiments (Section~\ref{sec:selected_baselines})
    \item Additional experimental results including pairwise registration with low overlap, indoor multi-view registration on 3DMatch and ScanNet, cross-domain multi-view registration benchmark, and offline SLAM pose estimation (Section~\ref{sec:additional_results})
    \item Additional information of the cross-domain multi-view registration benchmark (Section~\ref{sec:info_benchmark})
    \item Additional qualitative results and failure cases (Section~\ref{sec:additional_qualitative_results})
\end{itemize}

\section{Implementation Details}
\label{sec:implementation_details}

\subsection{Point cloud sampling and feature extraction}
For each point cloud $\mathbf{P}_i$, we first apply voxel downsampling with voxel size $v_d$ to obtain $\mathbf{P}_i^{v}$. We then apply statistical outlier removal to $\mathbf{P}_i^{v}$ to remove outliers. To ensure uniform sampling density across input point clouds, we determine the sample count $K_i$ proportionally to the spatial voxel coverage: we voxelize $\mathbf{P}_i^{v}$ with voxel size $v_c$ and let $V_i$ be the number of remaining points, then set $K_i = \lfloor \alpha_s V_i \rfloor$, where $\alpha_s$ is a hyperparameter controlling the FPS sampling density. We apply farthest point sampling (FPS) on $\mathbf{P}_i^{v}$ to sample $K_i$ points as the feature points $\mathbf{Q}_i = \{ \mathbf{q}_{i,k} \}_{k=1}^{K_i} \in \RR^{3 \times K_i}$. Typically, we got $K_i \in [200, 5000]$ for our training data.
For each feature point $\mathbf{q}_{i,k} \in \mathbf{Q}_i$, we extract a local patch by a ball query of radius $r_s = 20\,v_d$ in $\mathbf{P}_i^{v}$ and normalize the patch points. We then use the lightweight MiniSpinNet~\cite{ao2023cvpr-buffer, seo2025iccv} pretrained on 3DMatch~\cite{zeng2017cvpr} as our feature extractor $\mathcal{F}$ to compute a descriptor for each patch with maximum $512$ points and stack them into local features $\mathbf{F}_i \in \RR^{D \times K_i}$, where $D=32$.
During inference, there are only two tunable hyperparameters: the sampling density ratio $\alpha_s$ and the downsampling voxel size $v_d$. For ease of use, we set $\alpha_s = 0.2$ by default and determine $v_d$ adaptively based on the scene scale, requiring no scene-specific hyperparameter tuning. 

The pseudocode for this step is shown in \cref{alg:process-point-clouds}.

\begin{algorithm}[t]
    \LinesNumbered
    \caption{Point cloud sampling and miniSpinNet feature extraction}
    \label{alg:process-point-clouds}
    \KwIn{
      Set of input point clouds $\{\mathbf{P}_i\}$; \\
      voxel sizes $v_d$ (downsampling) and $v_c$ (coverage); \\
      FPS sampling ratio $\alpha_s$; patch radius $r_s$; \\
      miniSpinNet feature extractor $\mathcal{F}$.
    }
    \KwOut{Sampled points $\{\mathbf{Q}_i\}$ and features $\{\mathbf{F}_i\}$.}
    
    \BlankLine
    \ForEach{input point cloud $\mathbf{P}_i$}{
      Voxel-downsample $\mathbf{P}_i$ with voxel size $v_d$
      to obtain $\mathbf{P}_i^{v}$\;
      Apply statistical outlier removal to $\mathbf{P}_i^{v}$ to remove outliers\;
      Voxelize $\mathbf{P}_i^{v}$ with voxel size $v_c$
      and let $V_i$ be the number of remaining points\;
      Set $K_i \gets \lfloor \alpha_s V_i \rfloor$\;
      Apply FPS on $\mathbf{P}_i^{v}$ to sample $K_i$ points
      as feature points $\mathbf{Q}_i = \{ \mathbf{q}_{i,k} \}_{k=1}^{K_i}$\;
      
      \BlankLine
      For each $\mathbf{q}_{i,k} \in \mathbf{Q}_i$, extract a local patch by
      a ball query of radius $r_s$ in $\mathbf{P}_i^{v}$ and normalize the
      patch points\;
      Use $\mathcal{F}$ (miniSpinNet) to compute a descriptor
      for each patch and stack them into
      $\mathbf{F}_i \in \mathbb{R}^{D \times K_i}$\;
    }
    
    \BlankLine
    \Return $\{\mathbf{Q}_i\}$ and $\{\mathbf{F}_i\}$\;
    \end{algorithm}

To improve efficiency and reduce memory usage, we adopt the lightweight patch-wise network MiniSpinNet from BUFFER~\cite{ao2023cvpr-buffer} as our local feature descriptor.
MiniSpinNet is a compact SpinNet-style~\cite{ao2021cvpr-spinnet} network that encodes each input patch into a 32-dimensional feature descriptor by decreasing the voxelization hyperparameters and simplifying the 3D cylindrical convolution (3DCC) layers, making it nearly nine times faster than the vanilla SpinNet.

Currently, we use MiniSpinNet~\cite{ao2023cvpr-buffer} as our feature extractor. However, our framework is compatible with other feature extractors for encoding local geometry, including task-specific ones trained for registration (e.g., FCGF~\cite{choy2019iccv} and YOHO~\cite{wang2022mm-yoho}) as well as more general self-supervised backbones based on point transformers~\cite{wu2024cvpr-ptv3} (e.g., Sonata~\cite{wu2025cvpr-sonata} and Utonia~\cite{zhang2026arxiv-utonia}). We leave the exploration of other feature extractors as our future work.

\subsection{Flow matching model architecture}
Following RPF~\cite{sun2025neurips}, we use a diffusion transformer~\cite{peebles2023iccv} with alternating-attention blocks~\cite{wang2025cvpr-vggt} for conditional flow matching. 

Our transformer architecture comprises $L=10$ alternating-attention blocks with hidden dimension $d=512$ and $h=8$ attention heads, totaling 73 million parameters. The alternating-attention mechanism alternates between two types of attention layers: (i) per-view self-attention that operates within each point cloud to consolidate view-specific structure, and (ii) global attention over all point tokens across all views to fuse information and enable cross-view reasoning. This design allows the model to simultaneously capture local geometric structure within each view and global relationships across multiple views.

The model's input consists of the noisy point cloud $\mathbf{X}(t)$ at time step $t$ and a conditioning signal $\mathbf{C} = f_\text{emb}(\bar{\mathcal{Q}}, \mathcal{F})$ obtained via a linear feature embedder $f_\text{emb}$. The conditioning $\mathbf{C}$ concatenates three components: (i) local geometric descriptors $\mathcal{F}$ extracted by MiniSpinNet for each sampled keypoint, (ii) positional encodings of the normalized point coordinates $\bar{\mathcal{Q}}$ using a multi-frequency Fourier feature mapping~\cite{mildenhall2020eccv}. The flow matching network $\mathbf{V}_{\theta}$ takes $\mathbf{X}(t)$ and $\mathbf{C}$ as input and predicts the velocity field $\nabla_t \mathbf{X}(t)$ that transports the noisy points toward the target registered configuration. 

Unlike RPF~\cite{sun2025neurips}, we do not take the point-wise normals and the scalar view index as additional conditioning signals. We also do not rely on a PTv3-based encoder~\cite{wu2024cvpr-ptv3} pretrained for the overlapping prediction task. These design choices make our model simpler and can be applied to various cross-domain training datasets.

\subsection{Flow matching model training}
As mentioned in the main paper, our flow matching model $\mathbf{V}_{\theta}$ is trained by minimizing the following conditional flow matching loss~\cite{lipman2023iclr}, which minimizes the difference between the predicted velocity field and the true velocity field:
\begin{equation}
    \mathcal{L}_{\text{FM}} = \mathbb{E}_{t, \mathbf{X}}\left[ \left\| \mathbf{V}_{\theta}(t,  \mathbf{X}(t) \mid \mathbf{C}) - \nabla_t \mathbf{X} (t)  \right\|^2 \right].
    \label{eq:fm}
\end{equation}

During the flow matching model training, we sample the time steps from a U-shaped distribution~\cite{lee2024neurips}.

We train our model using the Muon Optimizer~\cite{liu2025arxiv-muon}, with an initial learning rate of $5  \times 10^{-3}$ for matrix-like parameters and $5 \times 10^{-4}$ for vector-like parameters. From our experiments we find that using Muon instead of AdamW~\cite{kingma2015iclr} can achieve faster convergence and better performance. For the learning rate schedule, we halve the learning rate after 200, 240, 280, 320, 360, 400, and 500 epochs.
We train the model for three days with about 120k iterations using 32 NVIDIA A100 GPUs with 80\,GB VRAM each. The training data consists of registration instances with view counts ranging from 2 to 16. 

% Overall, training our flow-matching model requires on the order of $3\times 10^{20}$ floating-point operations (FLOPs).

\subsection{Training data curation}

We curate both the scan-level and submap-level training samples using the same script with different settings.

For each dataset, given the per-frame poses, we first select keyframes based on temporal and spatial thresholds, which removes redundant frames when the sensor is stationary or moving slowly. 
For datasets lacking accurate and globally consistent reference poses, we use a state-of-the-art SLAM system~\cite{pan2024tro} to estimate the poses for data curation.
For most LiDAR-based datasets (e.g. KITTI is already deskewed), we additionally apply deskewing (motion undistortion) to the keyframe point clouds when point-wise timestamps are available. For each sequence with $M$ keyframes, we aim to generate $N_{\text{target}} = \beta M$ training samples, where $\beta$ controls the number of samples per keyframe. For every sample, we randomly select $N \in [N_{\text{min}}, N_{\text{max}}]$ point clouds. Each point cloud is constructed by accumulating points from $F \in [F_{\text{min}}, F_{\text{max}}]$ consecutive keyframes, and the resulting point clouds do not share frames. We then transform the point clouds to the world frame using the corresponding keyframe poses. For each sample, we allow at most $T_{\max}$ attempts to find a valid configuration. A sample is considered valid only if (i) all point clouds are spatially close to each other, i.e. the pairwise distances between their centers are below a threshold $d_{\text{max}}$, and (ii) the point clouds are not isolated from each other, i.e. they form a connected graph under a minimum overlap-ratio threshold $\epsilon_{\text{overlap}}$. 
The overlap ratio between two point clouds is computed as the ratio of the number of occupied voxels in their intersection to the number of occupied voxels in their union, evaluated on a voxel grid with an adaptively set voxel size $v_{\text{overlap}}$. We set a very small overlapping ratio threshold $\epsilon_{\text{overlap}}$ (0.5\%-2\%) to add some hard samples that allow the model to learn to register low-overlapped point clouds.
Whenever we find a valid set of point clouds in an attempt, we save it as a training sample. We set $N_{\text{min}} = 2$ and $F_{\text{min}} = 1$ for all datasets. For scan-based samples, we set $F_{\text{max}} = 1$, and for submap-based samples, we use $F_{\text{max}} > 1$.

The pseudocode for generating the training samples is shown in \cref{alg:process-sequence}.

\begin{algorithm}[!htbp]
  \SetAlFnt{\footnotesize}
  \LinesNumbered
  \caption{Generate training samples from a sequence}
  \label{alg:process-sequence}
  \KwIn{
    Per-frame poses $\{\mathbf{T}_i\}$, point cloud scans $\{\mathbf{S}_i\}$; \\
    Keyframe thresholds $(\tau_{\text{time}}, \tau_{\text{space}})$; \\
    Sampling parameters: $\beta$, $T_{\max}$, $N_{\min}, N_{\max}$, $F_{\min}, F_{\max}$; \\
    Spatial and overlap thresholds: $d_{\max}$, $\epsilon_{\text{overlap}}$, $v_{\text{overlap}}$; \\
  }
  \KwOut{Generated training samples for training.}
  
  \BlankLine
  Select keyframe indices $\mathcal{K} \gets \textsc{SelectKeyframes}(\{\mathbf{T}_i\},
  \tau_{\text{time}}, \tau_{\text{space}})$\;
  \If{deskewing enabled}{
    \ForEach{$k \in \mathcal{K}$}{
      Apply $\textsc{Deskew}$ to $\mathbf{S}_k$ if pointwise timestamps are available\;
    }
  }
  
  \BlankLine
  Let $M \gets |\mathcal{K}|$, $N_{\text{target}} \gets \beta M$\;
  
  \BlankLine
  \For{$n = 1$ \KwTo $N_{\text{target}}$}{
    \For{$t = 1$ \KwTo $T_{\max}$}{
      Sample $N \sim \mathcal{U}\{N_{\min}, N_{\max}\}$\;
      Sample $N$ disjoint keyframe intervals
      $\{I_j\}_{j=1}^N$ from $\mathcal{K}$ with lengths
      $F_j \sim \mathcal{U}\{F_{\min}, F_{\max}\}$\;
      
      \BlankLine
      Initialize $\mathcal{M} \gets [\,]$, $\mathcal{C} \gets [\,]$\;
      \For{$j = 1$ \KwTo $N$}{
        Accumulate a point cloud
        $\mathbf{M}_j \gets \textsc{AccumulateFrames}(\{\mathbf{S}_i\}_{i \in I_j})$\;
        Transform $\mathbf{M}_j$ to the world frame using
        $\{\mathbf{T}_i\}_{i \in I_j}$\;
        Compute point cloud center $\mathbf{c}_j$ from $\mathbf{M}_j$ and append to $\mathcal{C}$\;
        Append $\mathbf{M}_j$ to $\mathcal{M}$\;
      }
      
      \BlankLine
      \If{$\exists (a,b)$ such that $\|\mathbf{c}_a - \mathbf{c}_b\|_2 > d_{\max}$}{
        \textbf{continue} to next attempt\;
      }
      
      \BlankLine
      Build a graph $G$ over nodes $\{1, \dots, N\}$ with edge
      $(a,b)$ if
      $\textsc{OverlapRatio}(\mathbf{M}_a, \mathbf{M}_b, v_{\text{overlap}})
      \ge \epsilon_{\text{overlap}}$\;
      \If{$G$ is connected}{
        \textsc{SaveTrainingSample}($\mathcal{M}$)\;
        \textbf{break}\;
      }
    }
  }
\end{algorithm}

\begin{table*}[!htbp]
    \centering
    \footnotesize
    \caption{Summary of the training datasets with parameter settings for data curation. Sampling parameters: $N_{\max}$, $F_{\max}$; Spatial and overlap thresholds: $d_{\max}$, $\epsilon_{\text{overlap}}$; Point cloud preprocessing parameters: $\alpha_s$, $v_d$.}
    \label{tab:training_dataset_params}
    \resizebox{\textwidth}{!}{%
    \begin{tabular}{lcccccccccccc}
      \toprule
      \multirow{2}{*}{Dataset} & \multirow{2}{*}{Scenario} & \multirow{2}{*}{Sensor} & \multirow{2}{*}{\# Scenes} & \multirow{2}{*}{Type} & \multirow{2}{*}{\# Samples} & \multirow{2}{*}{\# P. Clouds} & \multicolumn{2}{c}{Sampling Parameters} & \multicolumn{2}{c}{Spatial Thre.} & \multicolumn{2}{c}{Preprocess} \\
      \cmidrule(lr){8-9} \cmidrule(lr){10-11} \cmidrule(lr){12-13}
      & & & & & & & $N_{\max}$ & $F_{\max}$ & $d_{\max}$ & $\epsilon_{\text{overlap}}$ & $\alpha_s$ & $v_d$ \\
      \midrule
      \rowcolor{gray!20}
      \multicolumn{13}{l}{\textit{Outdoor LiDAR}} \\
      \addlinespace[0.5em]
      \multirow{2}{*}{KITTI~\cite{geiger2012cvpr}} & \multirow{2}{*}{Germany; urban \& highway} & \multirow{2}{*}{Velodyne-64} & \multirow{2}{*}{22} & Scan & 1,226 & 2,852 & 8 & 1 & 100.0 & 1\% & 0.2 & 0.25 \\
      & & & & Submap & 3,810 & 16,453 & 10 & 600 & 400.0 & 0.5\% & 0.05 & 0.25 \\
      \addlinespace[1em]
      \multirow{2}{*}{KITTI360~\cite{liao2022pami}} & \multirow{2}{*}{Germany; urban} & \multirow{2}{*}{Velodyne-64} & \multirow{2}{*}{9} & Scan & 3,002 & 6,223 & 8 & 1& 100.0& 1\% & 0.2& 0.25 \\
      & & & & Submap & 7,530 & 21,255 & 10 & 600 & 400.0 & 0.5\% & 0.05 & 0.25 \\
      \addlinespace[1em]
      \multirow{2}{*}{Apollo~\cite{huang2018cvprws}} & \multirow{2}{*}{USA; urban \& highway} & \multirow{2}{*}{Velodyne-64} & \multirow{2}{*}{11} & Scan & 3,343 & 7,874 & 8 & 1 & 100.0 & 1\% & 0.2 & 0.25 \\
      & & & & Submap & 6,660 & 25,036 & 10 & 600 & 400.0 & 0.5\% & 0.05 & 0.25 \\
      \addlinespace[1em]
      \multirow{2}{*}{MulRAN~\cite{kim2020icra}} & \multirow{2}{*}{South Korea; urban \& campus} & \multirow{2}{*}{Ouster-64} & \multirow{2}{*}{4} & Scan & 898 & 1900 & 8 & 1 & 100.0 & 2\% & 0.2 & 0.25 \\
      & & & & Submap & 1,388 & 4,477 & 10 & 600 & 400.0 & 0.5\% & 0.05 & 0.25 \\
      \addlinespace[1em]
      \multirow{2}{*}{Oxford Spires~\cite{tao2025ijrr-oxfordspires}} & \multirow{2}{*}{UK; campus} & \multirow{2}{*}{Hesai-64} & \multirow{2}{*}{6} & Scan & 1,356 & 5,781 & 10 & 1 & 60.0 & 2\% & 0.2 & 0.25 \\
      & & & & Submap & 541 & 2,763 & 10 & 200 & 150.0 & 1\% & 0.1 & 0.25 \\
      \addlinespace[1em]
      \multirow{2}{*}{VBR~\cite{brizi2024icra-vbr}} & \multirow{2}{*}{Italy; urban \& campus} & \multirow{2}{*}{Ouster-64} & \multirow{2}{*}{5} & Scan & 3,371 & 8,647 & 8 & 1 & 80.0 & 1\% & 0.2 & 0.25 \\
      & & & & Submap & 1,906& 8,511& 10 & 500 & 300.0 & 0.5\% & 0.05 & 0.25 \\
      \addlinespace[1em]
      \multirow{2}{*}{UrbanNav~\cite{hsu2023navi-urbannav}} & \multirow{2}{*}{China; urban} & \multirow{2}{*}{Velodyne-32} & \multirow{2}{*}{4} & Scan & 1,912& 4,228& 8 & 1 & 100.0 & 1\% & 0.2 & 0.25 \\
      & & & & Submap & 979 & 3,540 & 10 & 600 & 400.0 & 0.5\% & 0.05 & 0.25 \\
      \addlinespace[1em]
      \multirow{2}{*}{HeLiPR~\cite{jung2024ijrr}} & \multirow{2}{*}{South Korea; urban} & \multirow{2}{*}{Ouster-128, Avia, Aeva} & \multirow{2}{*}{3} & Scan & 1,808& 3,691 & 8 & 1 & 100.0 & 1\% & 0.2 & 0.25 \\
      & & & & Submap & 3,624& 10,882& 10 & 600 & 400.0 & 0.5\% & 0.05 & 0.25 \\
      \addlinespace[1em]
      \multirow{2}{*}{Boreas~\cite{burnett2023ijrr-boreas}} & \multirow{2}{*}{Canada; urban} & \multirow{2}{*}{Velodyne-128} & \multirow{2}{*}{2} & Scan & 1,131 & 2,311 & 5& 1& 100.0 & 1\% & 0.2 & 0.25 \\
      & & & & Submap & 1,429& 3,613& 10 & 600 & 400.0 & 0.5\% & 0.05 & 0.25 \\
      \addlinespace[1em]
      WildPlace~\cite{knights2023icra-wildplaces} & Australia; forest & Velodyne-16 & 2 & Submap & 1,167& 2,613& 5& 600& 300.0 & 1\% & 0.1 & 0.25 \\
      \addlinespace[1em]
      NuScenes~\cite{caesar2020cvpr} & USA \& Singapore; urban & Velodyne-32 & 642 & Scan & 12,160 & 24,320 & 2 & 1 & 80.0 & 2\%& 0.5 & 0.25 \\
      \addlinespace[1em]
      KITTI-Carla~\cite{deschaud2021arxiv-kitticarla} & Synthetic; urban & Simulated-64 & 7 & Submap & 1,733& 7,729& 10& 600& 400.0& 0.5\%& 0.05& 0.25 \\
      \addlinespace[0.5em]
      \midrule
      \rowcolor{gray!20}
      \multicolumn{13}{l}{\textit{Indoor Depth Camera}} \\
      \addlinespace[0.5em]
      \multirow{2}{*}{3DMatch~\cite{zeng2017cvpr}} & \multirow{2}{*}{USA; office \& apartment} & \multirow{2}{*}{Kinect, RealSense, etc.} & \multirow{2}{*}{82} & Scan & 7,044& 14,088& 2& 1 & 10.0& 2\% & 1.0 & 0.02 \\
      & & & & Submap &3,904& 29,314& 16& 10 & 10.0& 1\% & 0.5 & 0.02\\
      \addlinespace[1em]
      ScanNet~\cite{dai2017cvpr} & USA; office \& apartment & Structure sensor & 661 & Submap & 10,840& 75,299 & 24& 50& 15.0& 1\%& 0.2& 0.02 \\
      \addlinespace[1em]
      ScanNet++~\cite{yeshwanth2023iccv-scannetpp} & Germany; office \& apartment & Faro, DSLR, iPhone & 220 & Scan & 9,306 & 101,743 & 24& 1 & 15.0& 1\%& 0.5& 0.02 \\
      \addlinespace[1em]
      NSS~\cite{sun2025jprs} & USA; office \& construction site & Matterport camera & 6 & Scan & 17,275& 72,866& 20& 1& 30.0& 0.1\% & 0.2& 0.05 \\
      \addlinespace[0.5em]
      \midrule
      \rowcolor{gray!20}
      \multicolumn{13}{l}{\textit{Object-centric}} \\
      \addlinespace[0.5em]
      ModelNet-40~\cite{wu2015cvpr} & Synthetic; CAD & - & (12,308) & Scan & 24,616 & 49,232 & 2 & 1 & - & - & - & 0.01 \\
      \addlinespace[0.5em]
      \midrule
      \rowcolor{gray!20}
      \textbf{Training Set} & & & \textbf{1,621} & & \textbf{118,143} & \textbf{457,195} & & & & & & \\
      \addlinespace[0.3em]
      \rowcolor{gray!20}
      \textbf{Total} & & & \textbf{1,685} & & \textbf{141,002} & \textbf{520,315} & & & & & & \\
      \bottomrule
    \end{tabular}%
    }
  \end{table*}

The details of the training datasets and used parameters for data curation ($N_{\max}$, $F_{\max}$, $d_{\max}$, $\epsilon_{\text{overlap}}$) and point cloud preprocessing ($\alpha_s$, $v_d$) are shown in \cref{tab:training_dataset_params}. In total, we curate 141k samples containing 520k point clouds, resulting in more than 10 billion points. The data covers a diverse range of scenes across 9 countries on 4 continents, captured by 9 types of LiDAR and 6 types of depth cameras with varying resolutions and scales. 

We split the data into training and validation sets. To ensure fair evaluation, we exclude sequences used for testing in commonly used benchmarks from the training set and designate them as validation sequences. For example, sequences 08--10 from the KITTI dataset are excluded from training. For some datasets not used in the testing benchmark (such as Nuscenes, Boreas, and WildPlace), we use all the sequences for training but keep 10\% randomly selected samples for the in-sequence validation.

\subsection{Pairwise registration testing data details}

The details of the six adopted testing datasets for pairwise registration evaluation and the evaluation success criteria are shown in \cref{tab:pairwise_testing_datasets}.

\begin{table*}[!htbp]
  \centering
  \footnotesize
  \caption{Summary of the testing datasets used for pairwise registration evaluation.}
  \label{tab:pairwise_testing_datasets}
  \resizebox{\textwidth}{!}{%
  \begin{tabular}{lccccccc}
    \toprule
    Dataset & Scenario & Sensor & Type & \# Samples & Scale [m] & Success Criteria  \\
    \midrule
    ModelNet~\cite{wu2015cvpr} & Synthetic; object & CAD & Object & 1,266 & 1 & TE 0.1m, RE 5$^\circ$ \\
    \addlinespace[0.3em]
    3DMatch~\cite{zeng2017cvpr} & USA; office \& apartment & Kinect, RealSense, etc. & Scan & 1,623 & 5 & Pointwise RMSE 0.2m \\
    \addlinespace[0.3em]
    3DLoMatch~\cite{huang2021cvpr-predator} & USA; office \& apartment & Kinect, RealSense, etc. & Scan & 1,781 & 5 & Pointwise RMSE 0.2m \\
    \addlinespace[0.3em]
    NSS~\cite{sun2025jprs} & USA; office \& construction site & Matterport camera & Scan & 1,125 & 10 & TE 0.2m, RE 10$^\circ$ \\
    \addlinespace[0.3em]
    ETH~\cite{pomerleau2012ijrr} & Switzerland; park & Hokuyo  & Scan & 713 & 100 & TE 2m, RE 5$^\circ$ \\
    \addlinespace[0.3em]
    KITTI~\cite{geiger2012cvpr} & Germany; urban \& highway & Velodyne-64 & Scan & 555 & 160 & TE 2m, RE 5$^\circ$ \\
    \bottomrule
  \end{tabular}%
  }
\end{table*}

\subsection{Evaluation metrics}
We evaluate pairwise registration performance using the registration success rate (\%), computed with thresholds on correspondence RMSE for 3DMatch and on translation and rotation errors for all other datasets; the exact thresholds are summarized in \cref{tab:pairwise_testing_datasets} and follow the settings of prior works~\cite{zeng2017cvpr, sun2025jprs, seo2025iccv}.

Given the ground-truth transformation $\mathbf{T}_{\text{gt}} = [\mathbf{R}_{\text{gt}} \mid \mathbf{t}_{\text{gt}}]$ and the estimated transformation $\mathbf{T}_{\text{est}} = [\mathbf{R}_{\text{est}} \mid \mathbf{t}_{\text{est}}]$, the translation error (TE) and rotation error (RE) are defined as:
\begin{align}
\text{TE} &= \|\mathbf{t}_{\text{gt}} - \mathbf{t}_{\text{est}}\|_2, \\
\text{RE} &= \arccos\left(\frac{\text{tr}(\mathbf{R}_{\text{gt}}^\top \mathbf{R}_{\text{est}}) - 1}{2}\right) \cdot \frac{180}{\pi},
\end{align}
where TE is in meters and RE is in degrees. For multi-view registration, we report the mean TE and RE across all scans in a sample.

For multi-view registration on the cross-domain benchmark, we adopt the \emph{edge success rate} and \emph{graph success rate} as the primary evaluation metrics. The edge success rate is the mean registration success rate over all valid edges (i.e., overlapping pairs) in a sample. The graph success rate is stricter: it requires every valid edge in the multi-view registration graph to be successful, thus measuring the overall consistency of the registration.
To handle diverse scene scales, we normalize the translation error by dividing by the longest axis of the bounding box of the ground-truth registered point cloud; a registration is considered successful if the normalized translation error is below 2.5\% and the rotation error is below 15\textdegree. 
We also report results under stricter thresholds (0.5\%, 3\textdegree) in this supplementary material.

For the additional multi-view registration experiments in this supplementary material, we adopt the following metrics: Chamfer distance (CD), normalized global RMSE, and the empirical cumulative distribution function (ECDF) of error.
The CD measures the bi-directional root-mean-squared distance between the registered point cloud and the ground-truth aggregated point cloud. Given the registered point cloud $\mathbf{P}_{\text{reg}}$ and the ground-truth point cloud $\mathbf{P}_{\text{gt}}$, the CD is computed as:
\begin{equation}
\small
\text{CD} = \sqrt{\frac{1}{2}\left(\frac{1}{|\mathbf{P}_{\text{reg}}|} \sum_{\mathbf{p} \in \mathbf{P}_{\text{reg}}} \min_{\mathbf{q} \in \mathbf{P}_{\text{gt}}} \|\mathbf{p} - \mathbf{q}\|_2^2 + \frac{1}{|\mathbf{P}_{\text{gt}}|} \sum_{\mathbf{q} \in \mathbf{P}_{\text{gt}}} \min_{\mathbf{p} \in \mathbf{P}_{\text{reg}}} \|\mathbf{q} - \mathbf{p}\|_2^2\right)}.
\end{equation}

The global RMSE is computed as the root-mean-squared error between the generated point cloud and the ground-truth registered point cloud using the known point-to-point correspondences, and is normalized by the longest axis of the ground-truth registered point cloud's bounding box to account for diverse scene scales.

The ECDF of error reports the fraction of samples whose error falls below a given threshold. Plotting the ECDF across thresholds summarizes the full error distribution and allows comparison of methods at different accuracy levels.

\section{Details on the Selected Baselines}
\label{sec:selected_baselines}

In the main paper, we did not include detailed descriptions of the baseline methods due to space limitations. Here we provide comprehensive descriptions and used training datasets of all baselines used in our experiments in \cref{tab:baselines}.

All pairwise registration baselines follow the standard correspondence matching and transformation estimation pipeline. All multi-view registration baselines except RPF~\cite{sun2025neurips} follow a two-stage pipeline: first performing pairwise registration along dense or sparse graph edges, then applying pose graph optimization to enforce global consistency. In contrast, our method is single-stage and directly generates the registered point cloud via flow matching, eliminating the need for exhaustive pairwise correspondence matching, transformation estimation, and pose graph optimization.

When evaluating on the cross-domain benchmark, we select models for learning-based baselines as follows.
For methods with multiple models trained on different datasets, we choose based on scene scale: ModelNet for object-centric, 3DMatch for indoor small-scale, and KITTI for outdoor large-scale scenes.
For methods with only a single model, we rescale the input to match the training scale at test time and then scale back for evaluation.

\begin{table*}[!htbp]
  \centering
  \footnotesize
  \caption{Description of baseline methods used in our experiments, organized by pairwise and multi-view registration tasks.}
  \label{tab:baselines}
  \resizebox{\textwidth}{!}{%
  {\setlength{\tabcolsep}{12pt}
  \begin{tabular}{lllp{10.6cm}}
    \toprule
    Method & Category & Venue & Description \\
    \midrule
    \rowcolor{gray!20}
    \multicolumn{4}{l}{\textit{Pairwise Registration Baselines}} \\
    \midrule
    \addlinespace[0.5em]
    FPFH~\cite{rusu2009icra} + FGR~\cite{zhou2016eccv} & Conventional & ECCV'16 & Handcrafted FPFH features combined with fast global registration via a robust cost function. \\
    \addlinespace[1em]
    FPFH~\cite{rusu2009icra} + TEASER~\cite{yang2020tro} & Conventional & TRO'20 & Handcrafted FPFH features with certifiably optimal pose estimation via truncated least squares and semidefinite relaxation, robust to high outlier rates. \\
    \addlinespace[1em]
    KISS-Matcher~\cite{lim2025icra-kissmatcher} & Conventional & ICRA'25 & Fast and robust registration combining efficient hand-crafted feature extraction with robust correspondence estimation, designed for scalability and generalization across diverse scenes. \\
    \addlinespace[1em]
    FCGF~\cite{choy2019iccv} & Deep learning & ICCV'19 & Sparse 3D convolutional network for dense geometric feature extraction, enabling dense correspondence matching for robust registration. Trained on 3DMatch or KITTI. \\
    \addlinespace[1em]
    Predator~\cite{huang2021cvpr-predator} & Deep learning & CVPR'21 & Overlap-aware network that predicts overlap scores and uses attention-weighted features to focus on overlapping regions, particularly effective in low-overlap scenarios. Trained on 3DMatch, KITTI or ModelNet. \\
    \addlinespace[1em]
    GeoTransformer~\cite{qin2022cvpr-geotransformer} & Deep learning & CVPR'22 & Keypoint-free method matching superpoints via transformation-invariant geometric features; uses optimal transport for dense correspondences without RANSAC. Trained on 3DMatch, KITTI or ModelNet.  \\
    \addlinespace[1em]
    BUFFER~\cite{ao2023cvpr-buffer} & Deep learning & CVPR'23 & Balances accuracy and efficiency via learned keypoint detection, patch feature embedding, and inlier correspondence generation. Trained on 3DMatch or KITTI. \\
    \addlinespace[1em]
    PARENet~\cite{yao2024iccv-parenet} & Deep learning & ECCV'24 & Rotation-equivariant, position-aware network for robust registration in low-overlap scenarios. Trained on 3DMatch or KITTI. \\
    \addlinespace[1em]
    BUFFER-X~\cite{seo2025iccv} & Deep learning & ICCV'25 & Extends BUFFER with improved zero-shot generalization via adaptive voxel sizing, farthest-point sampling, and patch-wise scale normalization. Originally trained on 3DMatch or KITTI; we additionally retrain a variant, denoted BUFFER-X$^*$, on our training data. \\
    \addlinespace[0.5em]
    \midrule
    \rowcolor{gray!20}
    \multicolumn{4}{l}{\textit{Multi-view Registration Baselines}} \\
    \midrule
    \addlinespace[0.5em]
    Pairwise registration + PGO~\cite{choi2015cvpr} & - & CVPR'15 & All-pair pairwise registration followed by a grow-based initialization and pose graph optimization, implemented by Open3D~\cite{zhou2018arxiv}.  \\
    \addlinespace[1em]
    SGHR~\cite{wang2023cvpr-sghr} & Deep learning & CVPR'23 & Constructs a sparse pose graph using NetVLAD-based overlap scores and conducts pairwise registration with YOHO~\cite{wang2022mm-yoho} on the edges from the sparse pose graph. Then it applies history-reweighted iterative reweighted least squares (IRLS) for stable, outlier-robust convergence. Trained on 3DMatch. \\
    \addlinespace[1em]
    RPF~\cite{sun2025neurips} & Deep learning & NeurIPS'25 & A flow-matching framework in Euclidean space that formulates  multi-part shape assembly as a conditional generative problem by learning a continuous point-wise velocity field that transports points to their assembled state. Trained on ModelNet and several object-centric shape assembly datasets. \\
    \addlinespace[0.5em]
    \bottomrule
  \end{tabular}}%
  }
  \vspace{-0.2cm}
\end{table*}

\section{Additional Experimental Results}
\label{sec:additional_results}

\subsection{Additional results on the pairwise registration with low overlap}

\begin{figure}[!htbp]
  \centering
  \includegraphics[width=\textwidth]{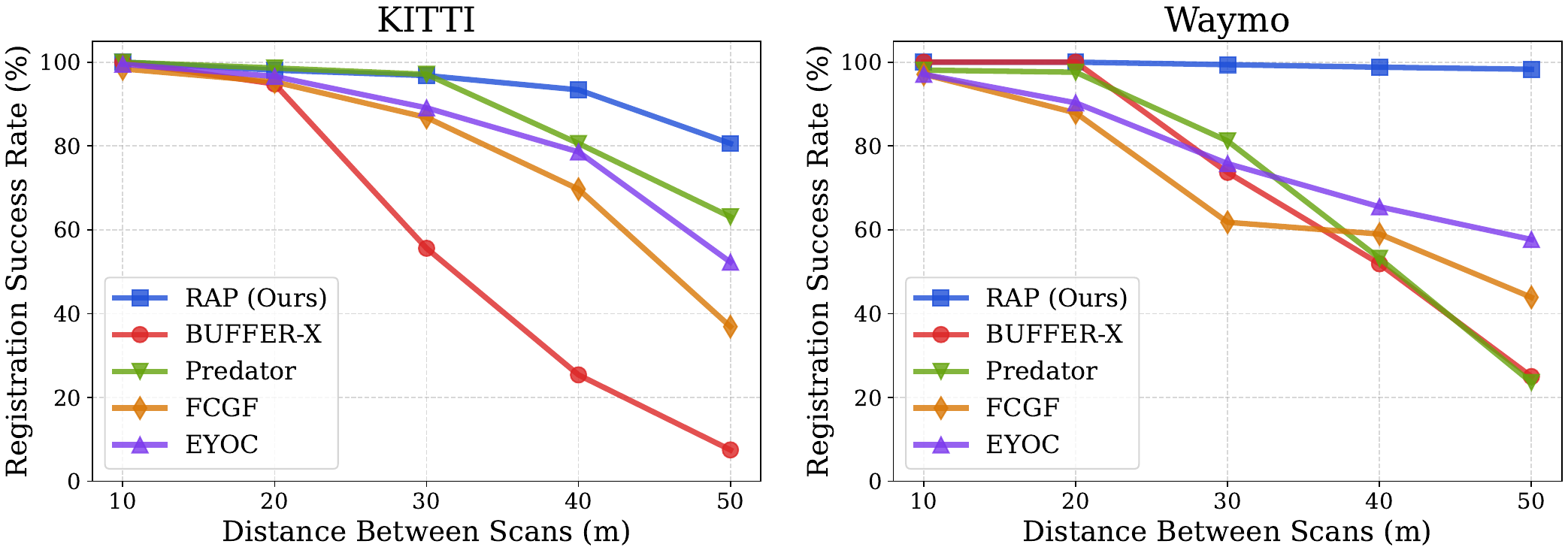}
  \vspace{-4pt}
  \caption{\textbf{Comparison of the pairwise point cloud registration:} Registration success rate with increasing spatial distance between the two point clouds (thus decreasing overlap ratio) on KITTI and Waymo datasets.}
  \label{fig:rr_dist}
  \vspace{-10pt}
\end{figure}

To further demonstrate our model's robustness to low overlap between point clouds, especially in the outdoor LiDAR scenarios, we follow EYOC~\cite{liu2024cvpr-eyoc} to curate testing data on both the KITTI~\cite{geiger2012cvpr} and Waymo~\cite{sun2020cvpr} with increasing spatial distance from 10\,m to 50\,m between point clouds (thus decreasing overlap). Note that KITTI at 10\,m is the setting used in the standard pairwise registration benchmark. As shown in \cref{fig:rr_dist}, our model shows superior performance over state-of-the-art methods such as BUFFER-X~\cite{seo2025iccv}, Predator~\cite{huang2021cvpr-predator}, FCGF~\cite{choy2019iccv}, and EYOC~\cite{liu2024cvpr-eyoc} with increasing scan distance.

\subsection{Additional results on indoor multi-view registration benchmark}

For multi-view registration evaluation, we also follow prior work~\cite{gojcic2020cvpr, wang2023cvpr-sghr} and use the common multi-view registration benchmarks on 3DMatch~\cite{zeng2017cvpr} and ScanNet~\cite{dai2017cvpr}. They are both indoor depth camera datasets.
We evaluate on sparse-view subsets (with $3 \leq N \leq 12$) of 3DMatch and ScanNet and report the translation error (TE), rotation error (RE), and Chamfer distance (CD). Comparisons with the baseline methods are shown in \cref{table:multiview_results}, demonstrating superior performance. The sparse-view setting makes the pairwise registration more challenging, thus most of the two-stage multi-view registration baselines fail to achieve good performance. 

\begin{table}[t]
  \centering
  \caption{Quantitative comparison of multi-view point cloud registration on 3DMatch and ScanNet dataset under the sparse view setting (with view count $3 \leq N \leq 12$). We report the mean rotation error (RE) in degrees, mean translation error (TE) in meters, and Chamfer distance (CD) in meters. Best results are shown in \textbf{bold}.}
  \vspace{-5pt}
{\scriptsize
\resizebox{0.7\linewidth}{!}{%
\begin{tabular}{l|ccc|ccc}
\toprule \midrule
& \multicolumn{3}{c|}{3DMatch} & \multicolumn{3}{c}{ScanNet} \\
\cmidrule(lr){2-4} \cmidrule(lr){5-7}
Method & RE $\downarrow$ & TE $\downarrow$ & CD $\downarrow$ & RE $\downarrow$ & TE $\downarrow$ & CD $\downarrow$ \\ \midrule
FGR~\cite{zhou2016eccv} + PGO & 52.81 & 0.71 & 0.49 & 68.80 & 1.43 & 0.76 \\
BUFFER-X~\cite{seo2025iccv} + PGO ~~ & 48.16 & 0.74 & 0.48 & 47.63 & 1.31 & 0.52 \\
LMVR~\cite{gojcic2020cvpr} & 15.46 & 0.44 & 0.16 & 46.10 & 0.87 & 0.50 \\
SGHR~\cite{wang2023cvpr-sghr} & 50.28 & 0.78 & 0.53 & 23.59 & 0.64 & 0.34 \\
\midrule
RAP (Ours) & \textbf{7.27} & \textbf{0.23} & \textbf{0.11} & \textbf{13.85} & \textbf{0.34} & \textbf{0.12} \\
\midrule \bottomrule
\end{tabular}%
}}
\label{table:multiview_results}
\vspace{-2pt}
\end{table}

\subsection{Additional results on cross-domain multi-view registration benchmark}

We show the comparison results on the cross-domain multi-view registration benchmark in \cref{tab:rapbench_eval_strict}, where we use the strict registration success rate (with a threshold of 0.5\% for the normalized translation error and 3$^\circ$ for the rotation error) instead of the standard threshold (2.5\% for the normalized translation error and 15$^\circ$ for the rotation error) used in the main paper.

\begin{table}[!t]
  \centering
  \caption{Comparison of the zero-shot testing performance for multi-view registration on the cross-domain multi-view registration benchmark. We report the \emph{strict} registration success rate (\%) calculated for edges and graphs (with a threshold of 0.5\% for the normalized translation error and 3$^\circ$ for the rotation error) in different scenarios as well as the average runtime. The best result is in \textbf{bold}, and the second best is \underline{underscored}. }
  \label{tab:rapbench_eval_strict}
  \vspace{-5pt}
  \resizebox{\linewidth}{!}{%
  {\scriptsize
  \begin{tabular}{l|c|cc|cc|cc|cc|cc|>{\columncolor{gray!10}}c>{\columncolor{gray!10}}c|c}
  \toprule
  \midrule
  \multicolumn{2}{c|}{\multirow{2}{*}{Method}} & \multicolumn{2}{c|}{Object} & \multicolumn{2}{c|}{Indoor} & \multicolumn{2}{c|}{Outdoor} & \multicolumn{2}{c|}{TLS} & \multicolumn{2}{c|}{Map} & \multicolumn{2}{c|}{\cellcolor{gray!10}All} & \multirow{2}{*}{Runtime (s)} \\
  \cmidrule(lr){3-4} \cmidrule(lr){5-6} \cmidrule(lr){7-8} \cmidrule(lr){9-10} \cmidrule(lr){11-12} \cmidrule(lr){13-14}
  \multicolumn{2}{c|}{} & edge & graph & edge & graph & edge & graph & edge & graph & edge & graph & \cellcolor{gray!10}edge & \cellcolor{gray!10}graph & \\
  \midrule
  FPFH+FGR~\cite{zhou2016eccv} & \multirow{7}{*}{\rotatebox[origin=c]{270}{+PGO}} & 9.7 & 2.1 & 1.2 & 0.4 & 2.2 & 0.5 & 2.7 & 0.6 & 4.3 & 1.6 & 2.5 & 0.7 & 19.7  \\
  FPFH+TEASER~\cite{yang2020tro} &  & 14.8 & 10.2 & 5.1 & 2.3 & 9.5 & 3.0 & 13.8 & 7.9 & 26.1 & 9.8 & 10.6 & 5.9 & 22.2 \\
  KISS-Matcher~\cite{lim2025icra-kissmatcher} &  & 16.4 & 13.6 & 15.9 & 5.7 &  21.8 & 7.4 & \underline{51.6} & 25.1 & \underline{75.1} & \underline{63.5} & 34.4 & 17.3 & 18.4 \\
  \cmidrule(lr){1-1} \cmidrule(lr){3-15}
  Predator~\cite{huang2021cvpr-predator} & & 1.7 & 0.8 & 4.4 & 2.1 & 3.8 & 1.4 & 2.5 & 1.1 & 5.0 & 2.2 & 3.4 & 1.5 & 51.8 \\
  BUFFER~\cite{ao2023cvpr-buffer} & & 1.3 & 0.5 & 15.3 & 6.7 & 8.2 & 3.7 & 16.3 & 6.3 & 1.4 & 0.3 & 12.0 & 5.4 & 250.3 \\
  BUFFER-X~\cite{seo2025iccv} & &  46.1 & 36.5 & 14.2 & 3.4 & 28.3 & 15.0 & 42.2 & 20.4 &11.5 & 2.9 & 28.9 & 13.7 & 36.0 \\
  BUFFER-X*~\cite{seo2025iccv} & & \underline{52.8} & \underline{40.1} & \underline{21.4} & 7.7 & \underline{36.9} & \underline{21.2} & 49.1 & \underline{25.8} & 17.3 & 6.6 & \underline{36.1} & \underline{20.8} & 36.0 \\
  \midrule
  \multicolumn{2}{l|}{SGHR~\cite{wang2023cvpr-sghr}} & 10.3 & 5.8 & 17.3 & \underline{8.6} & 14.5 & 7.6 & 19.8 & 12.0 & 12.8 & 5.8 & 16.2 & 9.1 & 49.7 \\
  \multicolumn{2}{l|}{RPF~\cite{sun2025neurips}} & 1.8 & 1.1 & 0.2 & 0.1 & 0.3 & 0.1 & 0.4 & 0.1 & 1.0 & 0.6 & 0.5 & 0.2 & \underline{11.5} \\
  \midrule
  \multicolumn{2}{l|}{RAP (Ours)}   & \textbf{74.7} & \textbf{58.1} & \textbf{60.2} & \textbf{36.5} & \textbf{84.0} & \textbf{76.6} &  \textbf{89.6} & \textbf{75.9} &  \textbf{92.7} & \textbf{90.4} & \textbf{79.8} & \textbf{65.8} & \textbf{8.9} \\
  \multicolumn{2}{l|}{RAP (Ours) w/o RF} & 68.8 & 51.4 & 53.0 & 29.9 & 81.3 & 70.5 & 84.6 & 68.3 &  89.6 & 86.8 & 74.9 & 59.0 & 8.9 \\
  \midrule
  \bottomrule
  \end{tabular}%
  }}
  \vspace{-4pt}
\end{table}

\begin{figure}[!htbp]
  \centering
  \includegraphics[width=1.0\linewidth]{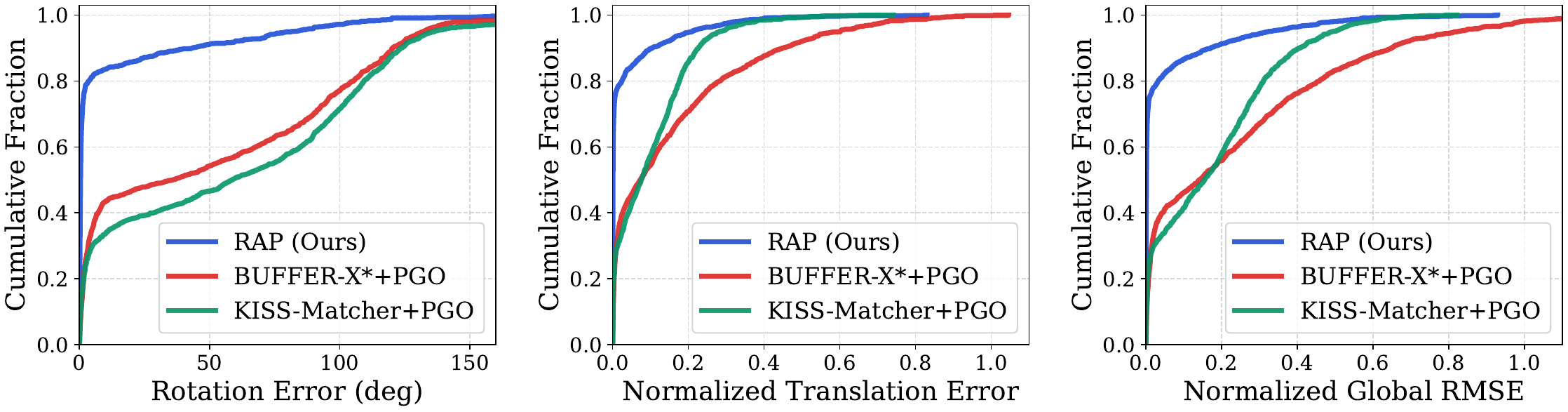}
    \vspace{-6pt}
    \caption{Comparison of the ECDF of the rotation error, normalized translation error, and normalized global RMSE on the cross-domain multi-view registration benchmark.}
  \label{fig:rapbench_ecdf_compare_all}
  \vspace{-8pt}
\end{figure}

\begin{figure}[!htbp]
  \centering
  \includegraphics[width=1.0\linewidth]{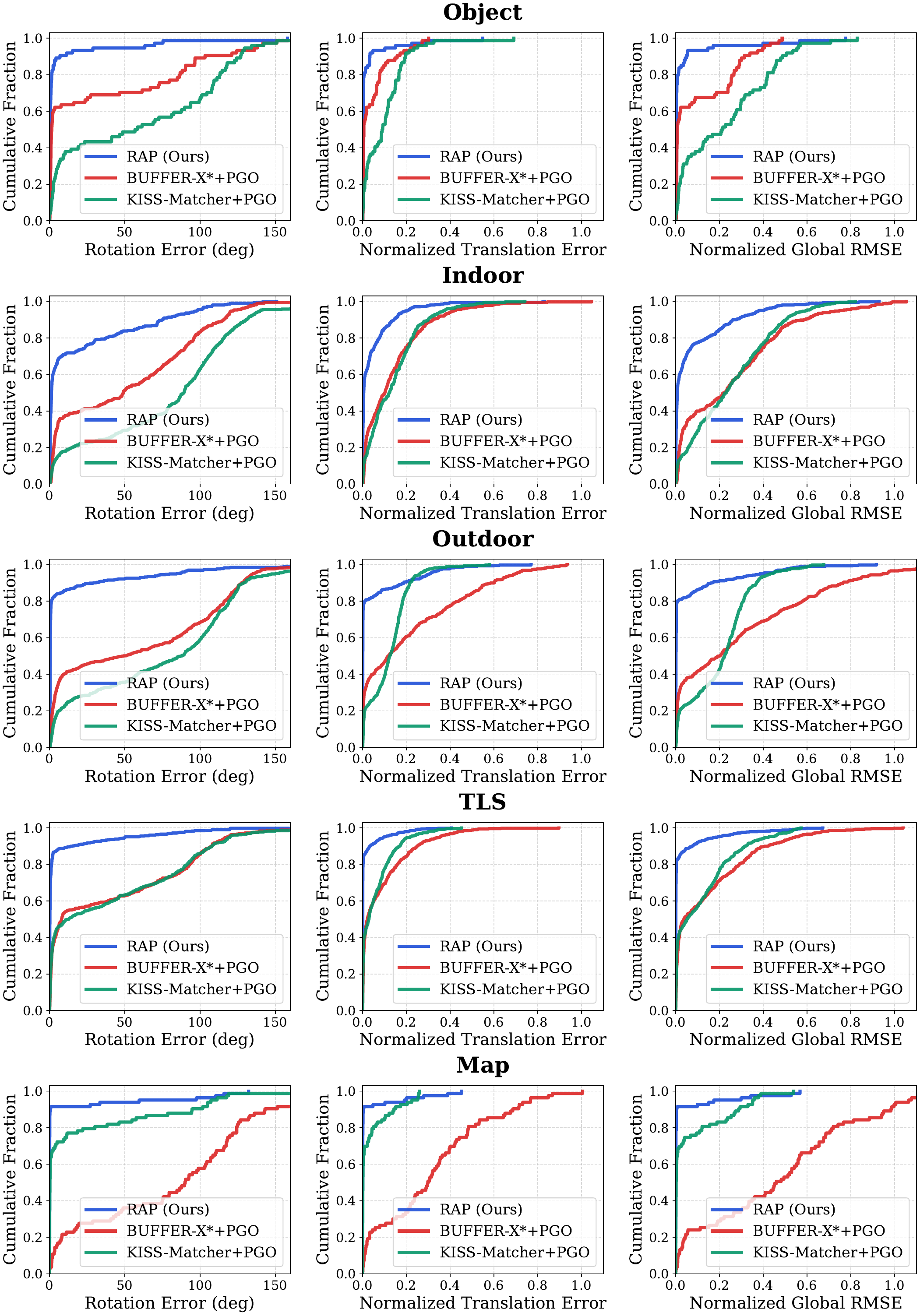}
    \vspace{-6pt}
    \caption{Comparison of the ECDF of the rotation error, normalized translation error, and normalized global RMSE on the five scenario categories of the cross-domain multi-view registration benchmark.}
  \label{fig:rapbench_ecdf_compare_categories}
  \vspace{-6pt}
\end{figure}

We additionally provide the ECDF plots for the rotation error, normalized translation error, and normalized global RMSE on the cross-domain multi-view registration benchmark in \cref{fig:rapbench_ecdf_compare_all} (overall) and \cref{fig:rapbench_ecdf_compare_categories} (by scenario category). We compare RAP with the two best-performing baselines: BUFFER-X$^*$+PGO and KISS-Matcher+PGO. The results show that RAP outperforms the baselines across all metrics and scenarios consistently.

\subsection{Pose estimation results for offline SLAM}

\begin{figure*}[!htbp]
  \centering
  \includegraphics[width=\linewidth]{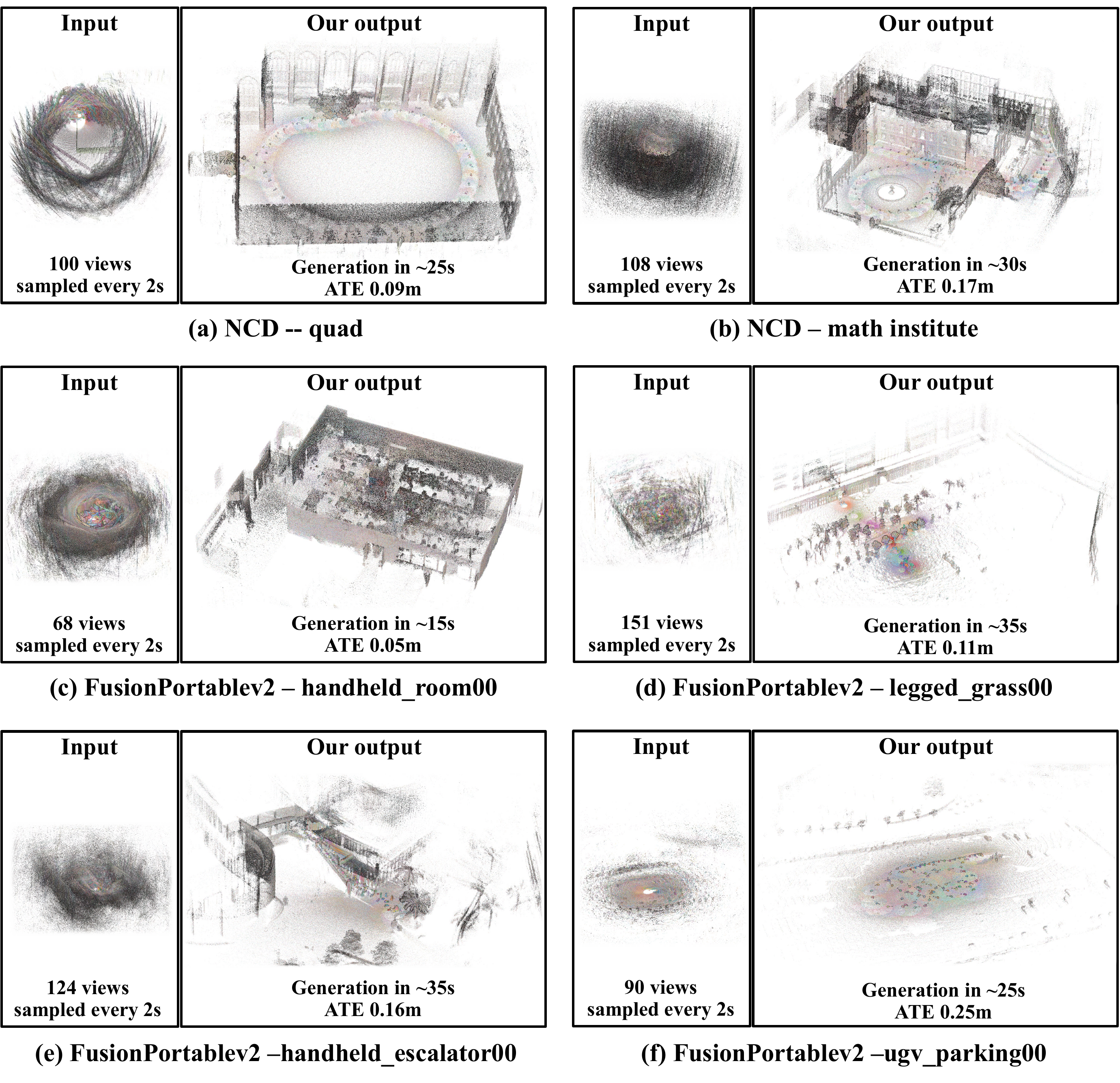}
  \vspace{-10pt}
  \caption{Offline SLAM results of our model on the FusionPortablev2~\cite{wei2025ijrr-fusionportablev2} and the Newer College dataset~\cite{ramezani2020iros}. Our model works zero-shot on these datasets, taking LiDAR point clouds per 2\,s (20 frames) in batches for evaluation. Different colors in the merged point cloud represent different LiDAR scans. We report the view count, generation runtime, and the localization ATE [m] for each sequence.}
  \label{fig:batch_slam_rap_fpv2_ncd}
  \vspace{-6pt}
\end{figure*}

\begin{figure*}[!htbp]
  \centering
  \begin{subfigure}[c]{0.43\linewidth}
    \centering
    \includegraphics[width=\linewidth]{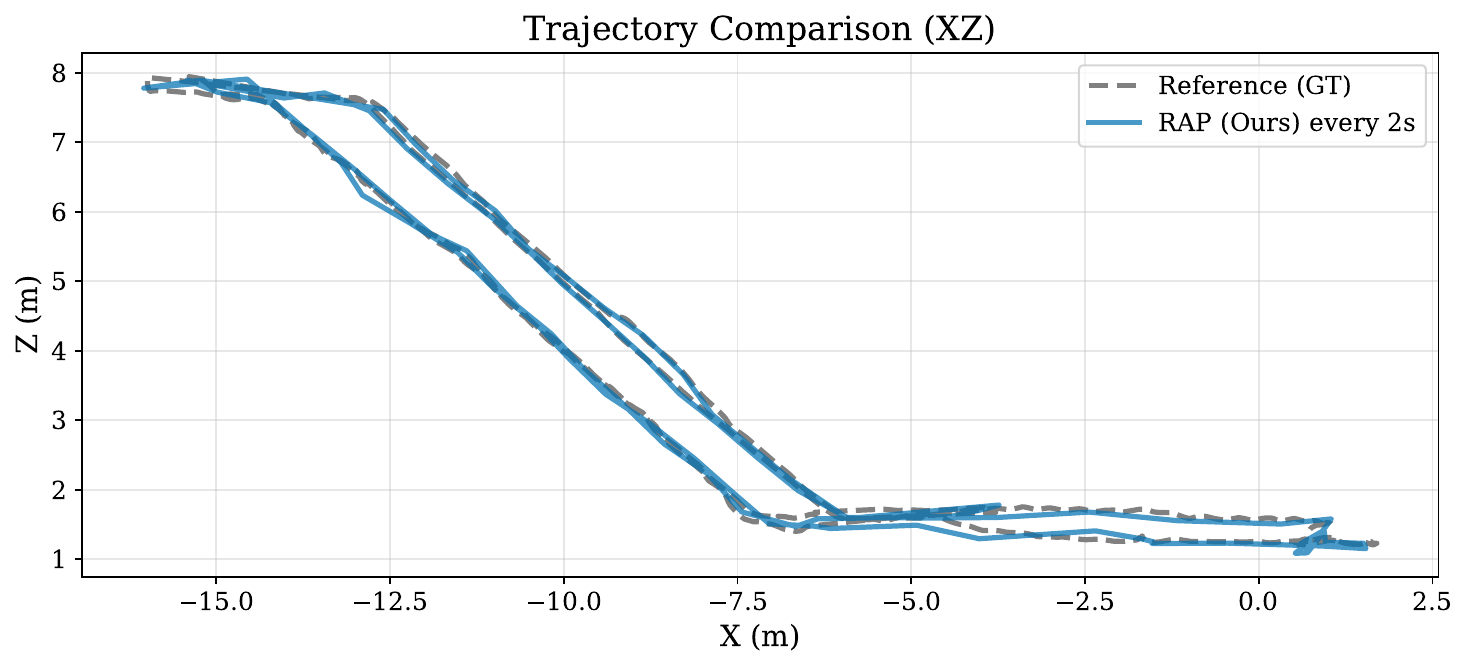}
    \caption{handheld\_escalator00 (XZ plane)}
  \end{subfigure}
  \hspace{0.02\linewidth}
  \begin{subfigure}[c]{0.38\linewidth}
    \centering
    \includegraphics[width=\linewidth]{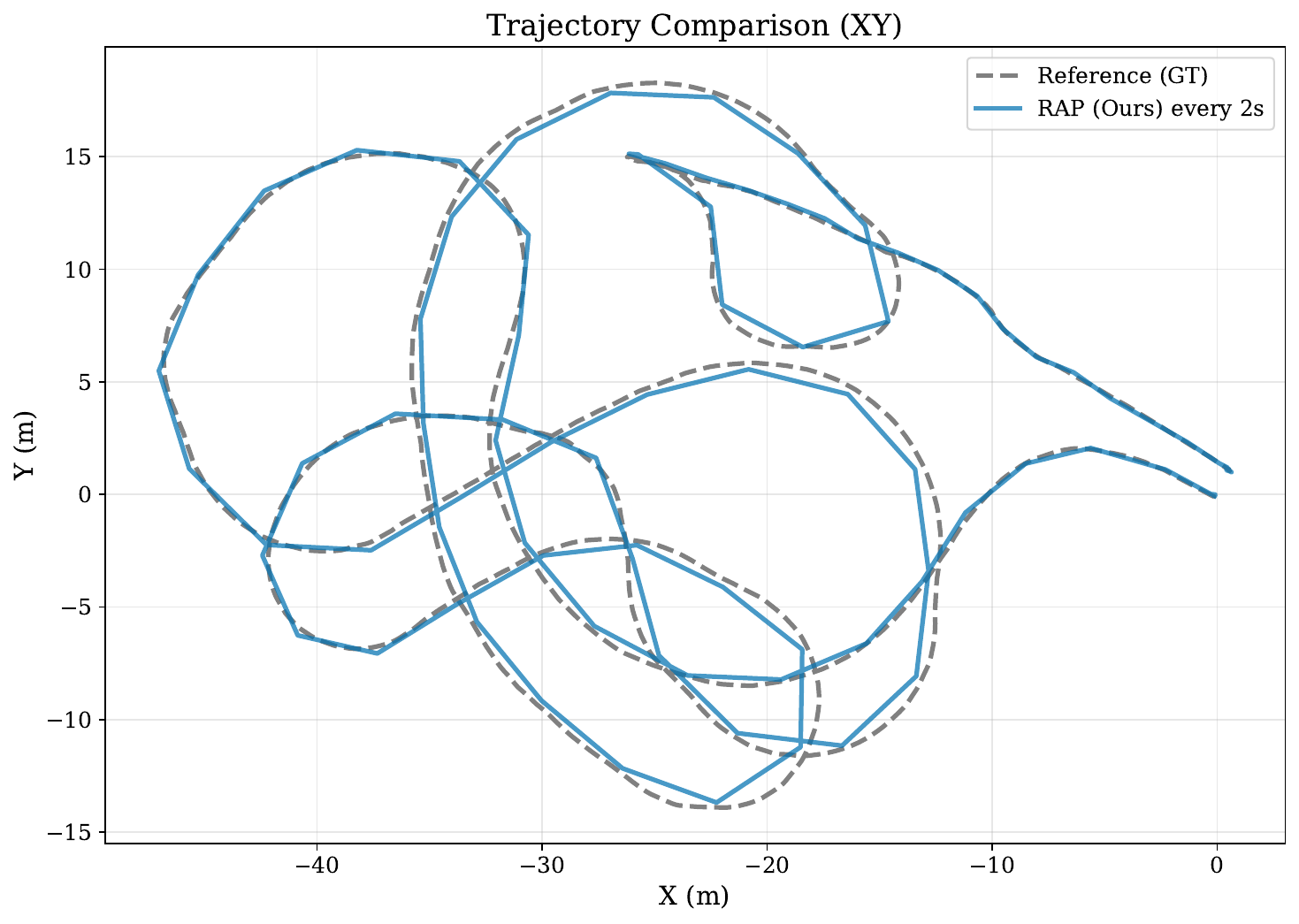}
    \caption{ugv\_parking00 (XY plane)}
  \end{subfigure}
  \caption{Estimated trajectories of RAP compared with the ground truth trajectories on the FusionPortablev2 dataset~\cite{wei2025ijrr-fusionportablev2}. Left: handheld\_escalator00 sequence shown in the XZ plane. Right: ugv\_parking00 sequence shown in the XY plane.}
  \label{fig:trajectory_ours}
  \vspace{-6pt}
\end{figure*}

One potential application of our model is offline SLAM, where we use it to estimate poses of LiDAR scans in a batch. We evaluate pose estimation accuracy using the absolute trajectory error (ATE) metric on the FusionPortablev2 dataset~\cite{wei2025ijrr-fusionportablev2}, which is an unseen dataset for our model. We compare our model with state-of-the-art SLAM systems: R3LIVE~\cite{jiarong2022icra} (LiDAR + camera + IMU), FAST-LIO2~\cite{wei2022tro} (LiDAR + IMU), VINS-Fusion~\cite{qin2018tro} (camera + IMU), and DROID-SLAM~\cite{teed2021neurips} (camera only) in \cref{tab:slam_ate_comparison}. We additionally show estimated trajectories of two sequences in \cref{fig:trajectory_ours} and merged point cloud maps on FusionPortablev2 and the Newer College dataset (NCD)~\cite{ramezani2020iros} in \cref{fig:batch_slam_rap_fpv2_ncd}.

Although our model is trained with at most 16 views, it can handle an arbitrary number of views during inference since we do not condition on view indices. However, feeding thousands of LiDAR scans at once is computationally expensive and may cause out-of-memory issues. We therefore sample LiDAR scans at 2\,s (20 frames) and 10\,s (100 frames) for evaluation, resulting in hundreds or tens of frames per sequence. Note that our model does not assume sequential order of the LiDAR scans and does not rely on any initial pose guess, which are typically required by conventional SLAM systems.

As shown in \cref{tab:slam_ate_comparison}, our model achieves comparable or better accuracy than the SLAM baselines (R3LIVE, FAST-LIO2, VINS-Fusion) that fuse LiDAR, camera, and IMU data, especially on scenes with geometric degenerations (e.g., ugv\_parking00). Intuitively, our model accomplishes an implicit LiDAR bundle adjustment, leading to better performance than approaches relying on incremental scan-to-map registration. The generation process takes about 30\,s for more than 100 sampled frames (and more than 2000 frames in the original sequence) on an NVIDIA A5000 GPU, which is comparable to or faster than the compared baselines that process at about 30\,ms per frame.

% Due to the large number of LiDAR frames in a sequence, 

\begin{table}[t]
  \centering
  \caption{SLAM localization accuracy comparison with the state-of-the-art SLAM systems on the FusionPortablev2 dataset~\cite{wei2025ijrr-fusionportablev2}. We calculate mean translation ATE [m] for each sequence. 
  The best result is shown in \textbf{bold}, and the second best result is \underline{underlined}. Our model works zero-shot on this dataset and takes the LiDAR point clouds per 2\,s (20 frames) and 10\,s (100 frames) in batches for evaluation.  VINS-Fusion w/ LC means VINS-Fusion with loop closure enabled.}
  \vspace{-5pt}
{\scriptsize
\resizebox{1.0\linewidth}{!}{%
  \begin{tabular}{lccc|ccccc}
  \toprule
  \midrule
  Method & LiDAR & Camera & IMU & \makecell{handheld\\room00} & \makecell{handheld\\escalator00} & \makecell{legged\\grass00} & \makecell{legged\\room00} & \makecell{ugv\\parking00} \\
  \midrule
  R3LIVE~\cite{jiarong2022icra}         & \checkmark & \checkmark & \checkmark & 0.057          & \underline{0.093} & \textbf{0.069} & \underline{0.068} & 0.424          \\
  FAST-LIO2~\cite{wei2022tro}          & \checkmark & $\times$   & \checkmark & 0.058          & \textbf{0.085} & 0.327          & 0.093          & 0.271          \\
  VINS-Fusion w/ LC~\cite{qin2018tro}  & $\times$   & \checkmark & \checkmark & 0.063          & 0.258          & 1.801          & 0.149          & 2.400          \\
  DROID-SLAM~\cite{teed2021neurips}    & $\times$   & \checkmark & $\times$   & 0.118          & 4.427          & 7.011          & 0.135          & 2.019          \\
  \midrule
  RAP (Ours) per 2\,s  & \checkmark & $\times$ & $\times$ & \textbf{0.050} & 0.164          & \underline{0.105} & 0.082          & \textbf{0.250} \\
  RAP (Ours) per 10\,s & \checkmark & $\times$ & $\times$ & \underline{0.052} & 0.140       & 0.108          & \textbf{0.064} & \underline{0.252} \\
  \midrule
  \bottomrule
  \end{tabular}%
}}
\label{tab:slam_ate_comparison}
\vspace{-2pt}
\end{table}

\section{Additional Information of the Cross-Domain Multi-View Registration Benchmark}
\label{sec:info_benchmark}

\cref{tab:benchmark_dataset_overview} provides an overview of the datasets used to compose the cross-domain multi-view registration benchmark separated by scenario category. Most of these datasets are originally not designed for registration evaluation.
The detailed statistics of the datasets can be found in the overview figure (\cref{fig:rapbench_info}).

\begin{table}[!htbp]
\centering
\caption{Overview of the datasets used to compose the cross-domain multi-view registration benchmark separated by scenario category.}
\vspace{-2pt}
\label{tab:benchmark_dataset_overview}
{\scriptsize
\resizebox{\linewidth}{!}{%
\setlength{\tabcolsep}{10pt}
\begin{tabular}{llll}
\toprule
Dataset & Location & Sensor & License \\
\midrule
\rowcolor{gray!20}
\multicolumn{4}{l}{\textit{Object}} \\
\midrule
C3VD~\cite{bobrow2023c3vd}                                   & -                            & clinical colonoscope                                                  & CC BY-NC-SA 3.0 \\
dSTORM nuclear lamina~\cite{vandelinde2011nprot-dstorm} & - & dSTORM microscope & - \\
Fruit shape completion~\cite{magistri2025icra}                & Germany                      & hand-held laser scanner, realsense RGB-D camera     & - \\
Stanford 3D Models~\cite{curless1996siggraph}  & USA                          & Cyberware scanner     & Custom research license \\
\midrule
\rowcolor{gray!20}
\multicolumn{4}{l}{\textit{Indoor Scan}} \\
\midrule
3DCSR~\cite{huang2021arxiv-comprehensive} & USA & Kinect RGB-D camera, 16-beam LiDAR, SfM & CC BY-NC-SA 3.0\\
Bonn RGB-D~\cite{palazzolo2019iros}   &  Germany & ASUS RGB-D camera & - \\
IILABS3D~\cite{ribeiro2025access}  & Portugal & Livox Mid-360, RoboSense Helios & CC BY-SA \\
Leg-KILO~\cite{ou2024ral} & China & Livox Mid-360 & CC BY 4.0 \\
Matterport3D~\cite{chang2017threedv-matterport3d}       & USA                          & Matterport RGB-D camera                           & Custom research license \\
TIERS~\cite{sier2023rs-tiers} & Finland & Velodyne 16-beam, Ouster 64/128-beam, Livox LiDAR & MIT \\
TUM RGB-D~\cite{sturm2012iros} & Germany & Kinect RGB-D camera & CC BY 4.0 \\
\midrule
\rowcolor{gray!20}
\multicolumn{4}{l}{\textit{Outdoor Scan}} \\
\midrule
Argoverse2~\cite{wilson2021neurips} & USA & 2 x LiDARs & CC BY-NC-SA 4.0 \\
Digiforest~\cite{malladi2025icra} & Switzerland & Hesai 32/64-beam LiDAR & - \\
MUST-C~\cite{chong2026sdata}  & Germany & Ouster 128 beam LiDAR, RIEGL scanner, LMI scanner & CC BY 4.0\\
NCLT~\cite{carlevaris-bianco2016ijrr}   & USA                      & Velodyne 32-beam LiDAR                                               & ODbL \\
Nuscenes~\cite{caesar2020cvpr}         & USA, Singapore            & Velodyne 32-beam LiDAR                                               & CC BY-NC 4.0 \\
PandaSet~\cite{xiao2021itsc-pandaset} & USA & Hesai 64-beam and foward-facing LiDAR & CC BY-NC-SA 4.0 \\
SGAB~\cite{chong2022sgab} & Singapore & 3 x Velodyne 16-beam LiDARs & Custom research license  \\
SubT~\cite{zhao2024cvpr-subt} & USA & Velodyne 16-beam LiDAR & - \\
Truckscenes~\cite{fent2024neurips-truckscenes}       & Germany                          & 6 x LiDARs                                               & CC BY-NC-SA 4.0\\
Waymo~\cite{sun2020cvpr}               & USA                          & 5 x LiDARs                                               & Custom research license \\
ZOD~\cite{alibeigi2023iccv-zod}       & 14 European countries                   & Velodyne 128-beam LiDAR                                         & CC BY-SA 4.0 \\
\midrule
\rowcolor{gray!20}
\multicolumn{4}{l}{\textit{TLS}} \\
\midrule
ETH3D~\cite{schops2017cvpr-eth3d}       & Switzerland                  & Leica TLS                                          & CC BY-NC-SA 4.0 \\
ETH-TLS~\cite{theiler2015jprs} & Switzerland & TLS & - \\
IndoorLRS~\cite{park2017iccv} &  -  & Faro TLS and Asus RGB-D camera & - \\
MCD~\cite{nguyen2024cvpr-mcd}          & Sweden, Germany, Singapore   & Leica and Faro TLS                                  & CC BY-NC-SA 4.0 \\
Oxford-TLS~\cite{tao2025ijrr-oxfordspires,ramezani2020iros} & UK & Leica TLS & - \\
R3DS~\cite{robotic3dscanrepository}                                   & Germany, Croatia                      & Riegl TLS                                                 & - \\
RESSO~\cite{chen2020tgrs-plade}   &  Netherlands, Saudi Arabia   & Leica and Faro TLS &  - \\
Semantic3D~\cite{hackel2017isprs} &  Switzerland & TLS & CC BY-NC-SA 3.0 \\
Tongji-Trees~\cite{wang2023jprs-globalmatch} & China & Z+F TLS & - \\
VMML~\cite{michailidis2019vc-aspire}   & Switzerland & Faro TLS & - \\
WHU-TLS~\cite{dong2020jprs-whutls} & China & RIEGL TLS & - \\
\midrule
\rowcolor{gray!20}
\multicolumn{4}{l}{\textit{Map}} \\
\midrule
Abenberg~\cite{hebel2013jprs}   & Germany & ALS & CC BY-NC-SA 4.0 \\
AHN~\cite{cserep2023jag-ahn}          & Netherlands  & ALS                                                 & CC0 \\
Alita~\cite{yin2022arxiv-alita}       & USA                          & Map built by Velodyne 16-beam LiDAR                         & BSD 3-Clause \\
Kimera-Multi~\cite{tian2023iros-kimeramultidata} & USA & Map built by Velodyne 16-beam LiDAR & MIT \\
MS-HKUSTGZ~\cite{hu2024arxiv-msmapping} & China & Map built by HESAI 32-beam LiDAR, Leica TLS & - \\
\bottomrule
\end{tabular}%
}}
\vspace{-4pt}
\end{table}

\section{Additional Qualitative Results}
\label{sec:additional_qualitative_results}

\begin{figure*}[!htbp]
\centering
\includegraphics[width=0.99\linewidth]{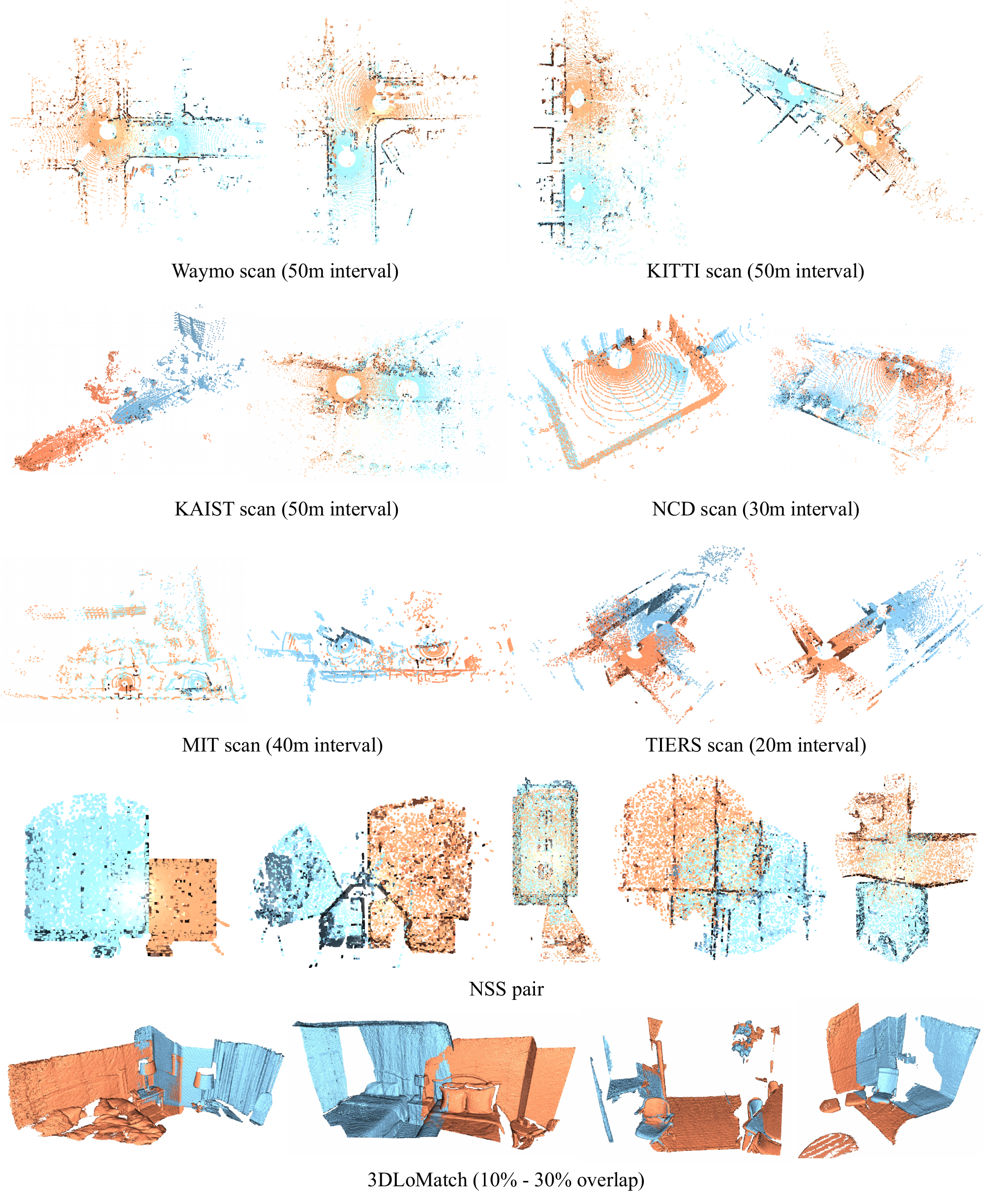}
\caption{Qualitative results of our model on pairwise registration with large scan interval (Waymo, KITTI, KAIST, NCD, MIT, TIERS) or low overlap ratio (NSS, 3DLoMatch).}
\label{fig:pairwise_benchmark_quali}
\end{figure*}

\begin{figure*}[!htbp]
  \centering
  \includegraphics[width=0.99\linewidth]{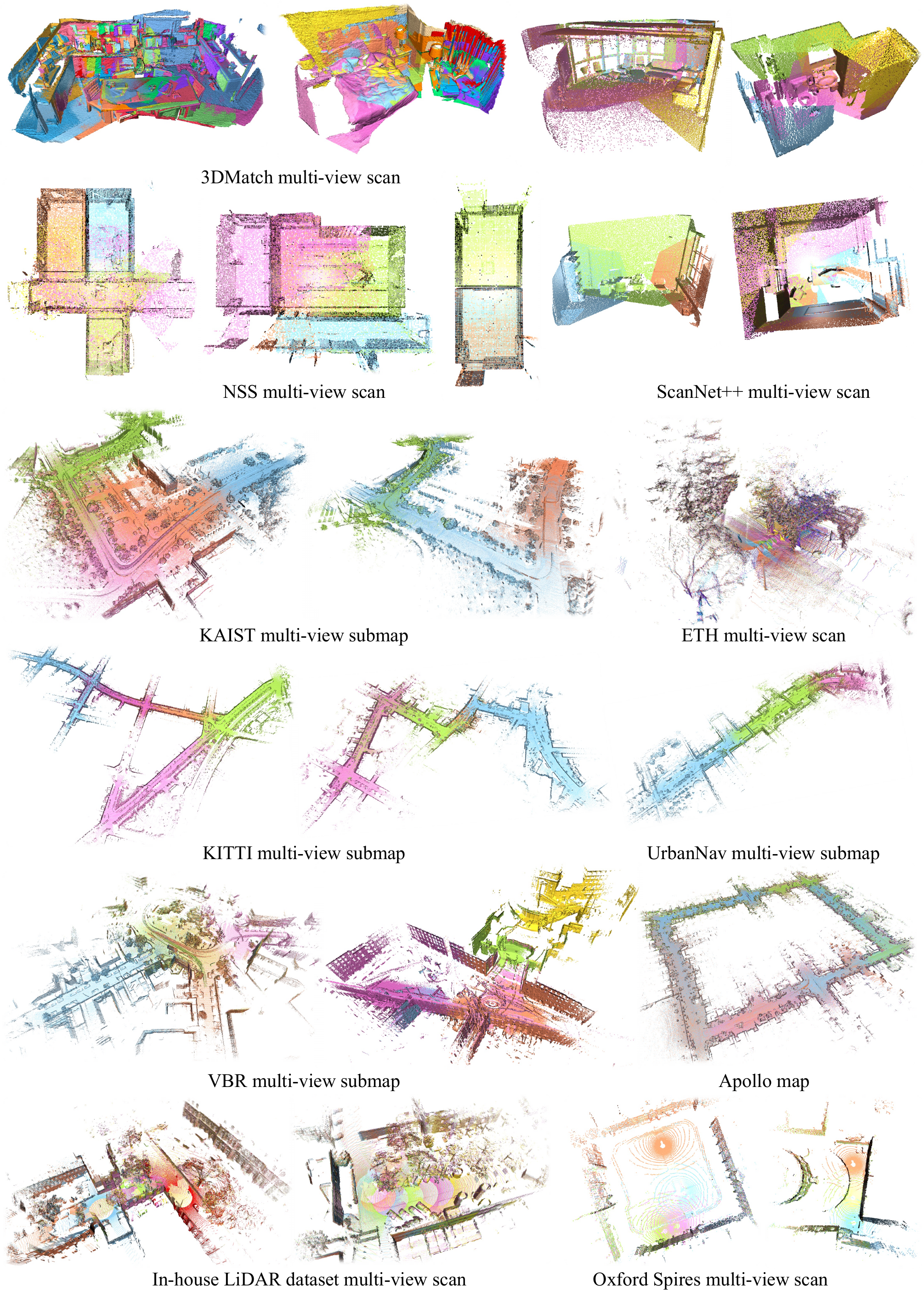}
  \caption{Qualitative results of our model on multi-view point cloud registration.}
  \label{fig:multiview_quali_1}
\end{figure*}

\begin{figure*}[!htbp]
  \centering
  \includegraphics[width=0.99\linewidth]{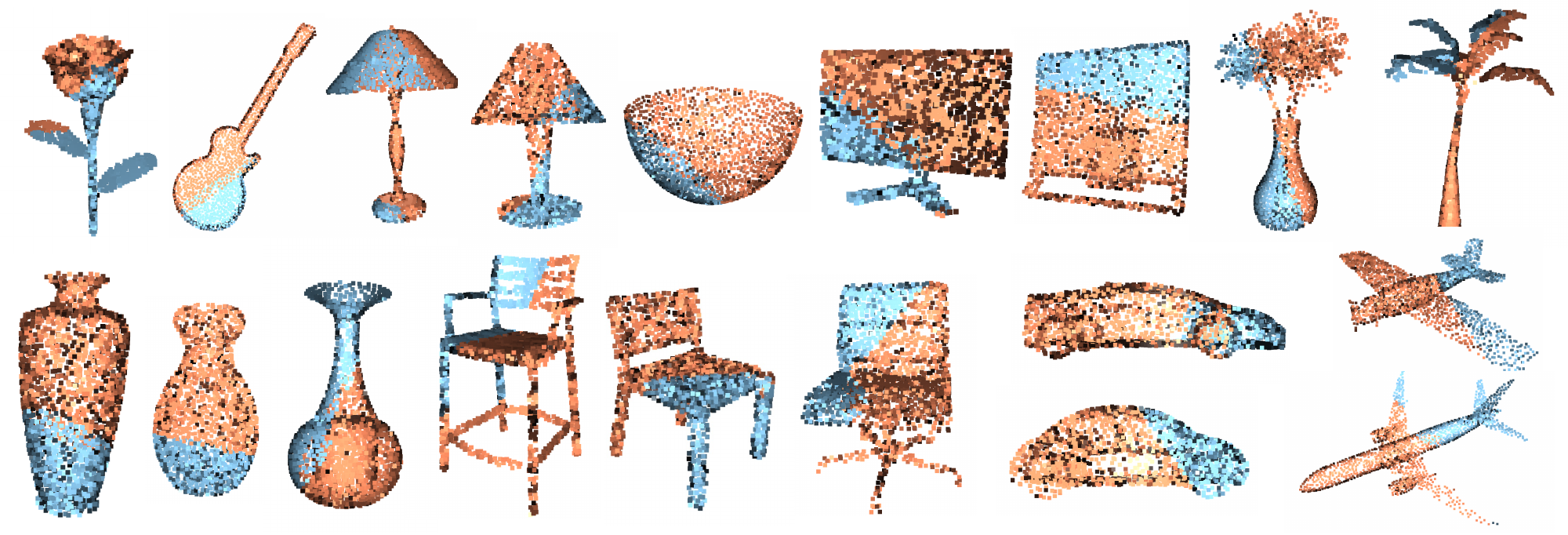}
  \caption{Qualitative results of our model on ModelNet for object-centric pairwise registration.}
  \label{fig:modelnet_quali}
\end{figure*}

\begin{figure*}[!htbp]
  \centering
  \includegraphics[width=0.99\linewidth]{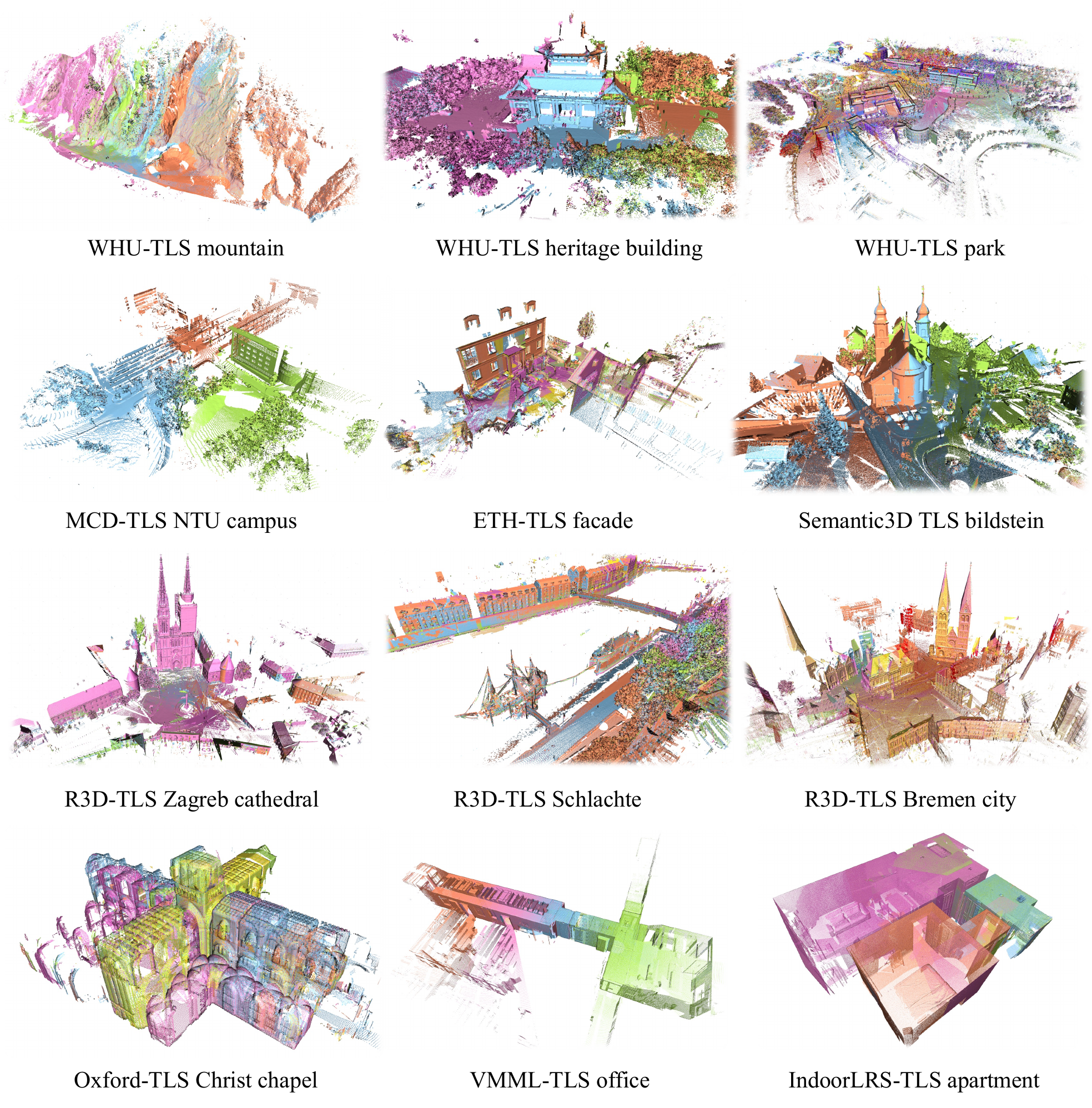}
  \caption{Qualitative results of our model on multi-view TLS point cloud registration.}
  \label{fig:multiview_quali_tls}
\end{figure*}

\begin{figure*}[!htbp]
  \centering
  \includegraphics[width=0.99\linewidth]{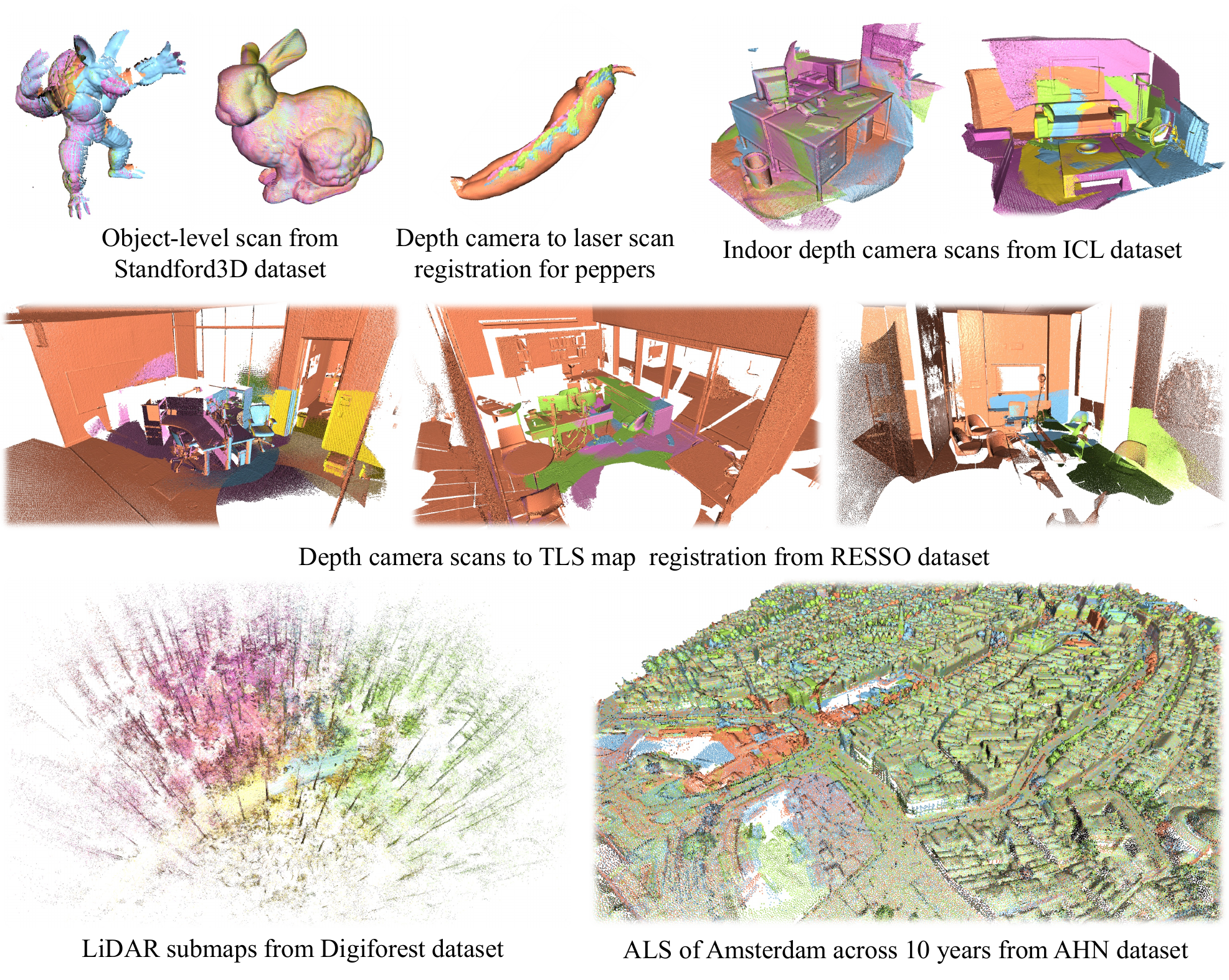}
  \caption{Qualitative results of our model on cross-domain multi-view registration benchmark, from object-centric to map-level scenes.}
  \label{fig:multiview_quali_crossdomain}
\end{figure*}

We provide additional qualitative results of our model on the pairwise and multi-view point cloud registration tasks on various datasets in this section, as shown in \cref{fig:pairwise_benchmark_quali}, \cref{fig:multiview_quali_1}, \cref{fig:multiview_quali_tls}, \cref{fig:multiview_quali_crossdomain}, and \cref{fig:modelnet_quali}.

\begin{figure*}[!htbp]
  \centering
  \includegraphics[width=0.99\linewidth]{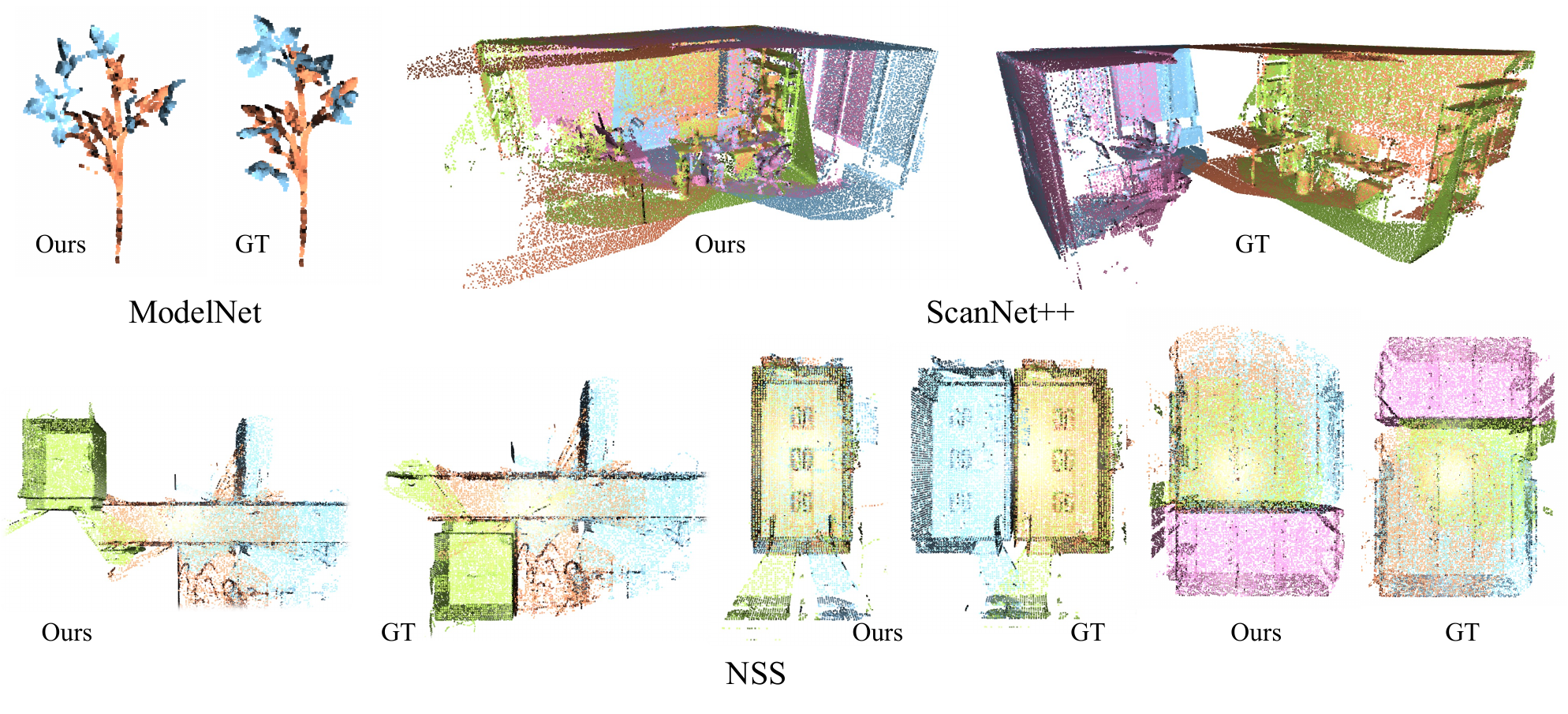}
  \caption{Failure cases of our model. We provide the prediction of our model and the ground truth for comparison.}
  \label{fig:failure_cases}
\end{figure*}

We also provide some failure cases of our model in \cref{fig:failure_cases}. It is shown that our model sometimes fails to correctly register the point clouds in the cases of very low overlap ratio or even no overlap due to the thickness of wall. From the examples shown for NSS dataset, it is shown that our model can still guess the reasonable global layout of the scene for the ambiguous failed cases.

% Flush all pending floats (figures, tables) before references
\clearpage
\begingroup
\renewcommand{\url}[1]{}
\bibliographystyle{splncs04}
\bibliography{glorified,new}
\endgroup
\end{document}